\documentclass[letterpaper]{article} 
\usepackage{aaai2026}  
\usepackage{times}  
\usepackage{helvet}  
\usepackage{courier}  
\usepackage[hyphens]{url}  
\usepackage{graphicx} 
\usepackage{natbib}  
\usepackage{caption} 
\usepackage{algorithm}
\usepackage{algorithmic}
\usepackage{booktabs}       
\usepackage{amsmath,amssymb}
\usepackage{multirow}
\usepackage{subcaption}
\usepackage{newfloat}
\usepackage{listings}

\usepackage{pifont}
\usepackage{amsmath,amssymb}

\usepackage{color}

\usepackage[table,xcdraw]{xcolor}
\definecolor{mygray}{gray}{0.88}

\def\1{\bm{1}}

\usepackage{fontawesome5}  
\usepackage{xcolor}

\usepackage{booktabs}
\usepackage{multirow}

\usepackage{mathrsfs}
\usepackage{makecell}

\def\rvepsilon{{\mathbf{\epsilon}}}

\def\rvx{{\mathbf{x}}}

\def\rvz{{\mathbf{z}}}

\def\rmA{{\mathbf{A}}}

\def\rmX{{\mathbf{X}}}

\usepackage{pifont}
\DeclareCaptionStyle{ruled}{labelfont=normalfont,labelsep=colon,strut=off} 
\floatstyle{ruled}
\newfloat{listing}{tb}{lst}{}
\floatname{listing}{Listing}
\usepackage{amsmath}
\usepackage{amssymb} 
\usepackage{array}
\usepackage{booktabs}
\usepackage{multirow}  
\usepackage{makecell}  
\usepackage{fixltx2e}
\usepackage{cite}
\usepackage{tikz-cd}
\usepackage{pgfplots}
\pgfplotsset{compat=1.17}
\usepackage{enumitem}
\usepackage{bibentry}

\usepackage{bibentry}

\usepackage{xcolor}
\newcommand{\lifeng}[1]
{{\color{black} #1}}

\usepackage{tabularx} 
\usepackage{booktabs}
\usepackage{subcaption}

\title{\lifeng{\textit{TSGDiff}: Rethinking Synthetic Time Series Generation\\from a Pure Graph Perspective}}
\author {
    Lifeng Shen\textsuperscript{\rm 1}\thanks{Corresponding Author.},
    Xuyang Li\textsuperscript{\rm 1},
    Lele Long\textsuperscript{\rm 1}
}
\affiliations {
    \textsuperscript{\rm 1}Chongqing Key Laboratory of Computational Intelligence, Key Laboratory of Cyberspace Big Data Intelligent Security, Ministry of Education, Sichuan-Chongqing Co-construction Key Laboratory of Digital Economy Intelligence and Key Laboratory of Big Data Intelligent Computing, Chongqing University of Posts and Telecommunications\\
    shenlf@cqupt.edu.cn, lixuyang.lee@foxmail.com, longlele0208@foxmail.com
}

\begin{document}
\maketitle

\begin{abstract}
\lifeng{
Diffusion models have shown great promise in data generation, yet generating time series data remains challenging due to the need to capture complex temporal dependencies and structural patterns. 
In this paper, we present \textit{TSGDiff}, a novel framework that rethinks time series generation from a graph-based perspective. Specifically, we represent time series as dynamic graphs, where edges are constructed based on Fourier spectrum characteristics and temporal dependencies. A graph neural network-based encoder-decoder architecture is employed to construct a latent space, enabling the diffusion process to model the structural representation distribution of time series effectively. Furthermore, we propose the Topological Structure Fidelity (Topo-FID) score, a graph-aware metric for assessing the structural similarity of time series graph representations. Topo-FID integrates two sub-metrics: Graph Edit Similarity, which quantifies differences in adjacency matrices, and Structural Entropy Similarity, which evaluates the entropy of node degree distributions. This comprehensive metric provides a more accurate assessment of structural fidelity in generated time series. 
Experiments on real-world datasets demonstrate that \textit{TSGDiff} generates high-quality synthetic time series data generation, faithfully preserving temporal dependencies and structural integrity, thereby advancing the field of synthetic time series generation. Source code and the extended version of the paper and  are available at https://github.com/jvaeylee/TSGDiff.
}
\end{abstract}

\section{Introduction}\label{sec:intro}
\lifeng{
Multivariate time series generation is of great importance in various domains, such as energy management \cite{r:10a}, financial market forecasting \cite{r:11a}, and medical monitoring \cite{r:12a}. 
Traditional generative models, including generative adversarial networks (GANs) \cite{r:13a} and variational autoencoders (VAEs) \cite{r:14a}, have made significant strides in this area. 
However, they still face notable challenges. 
First, these models often struggle to effectively capture the complex spatio-temporal dependencies among variables (e.g., TimeGAN \cite{r:19x} and TimeVAE \cite{r:1a}). 
Second, their reliance on Euclidean space assumptions limits their ability to represent the topological and structural characteristics inherent in time series data.

Effectively modeling temporal dependency structures is a key challenge in time series analysis. To address this, various techniques have been developed, including random historical skip connections \cite{kieu2019outlier}, multi-scale temporal links \cite{shen2021time}, and autoregressive temporal structures \cite{yu2016temporal}. These approaches are designed to capture intricate and long-range temporal relationships within time series data. 

In addition to capturing temporal dependencies, understanding and implementing generative mechanisms for time series data has garnered significant attention. Generative models such as TimeGAN \cite{r:19x}, TimeVAE \cite{r:1a}, and diffusion-based approaches like CSDI \cite{r:2a} and TimeDiff \cite{r:15a} 
have shown promise in synthesizing time series data by learning the underlying data distributions. For instance, diffusion-based models like Diffusion-TS \cite{r:24x} integrate seasonal-trend decomposition with denoising diffusion probabilistic models to learn temporal characteristics using Fourier-based loss terms. However, these methods often focus solely on raw data domains and fail to fully capture dynamic multivariate interactions. Additionally, architectures such as the encoder-decoder transformer in Diffusion-TS process spatial and temporal information separately, which limits their ability to model complex interdependencies between variables.

More recently, graph-based methods \cite{r:4a,r:5a,cheng2020time2graph} have emerged as a powerful alternative in spatiotemporal prediction tasks. 
For example, STGCN \cite{r:4a} combines graph and temporal convolutions to model spatio-temporal patterns, while FourierGNN \cite{r:5a} uses spectral graph theory and Fourier transforms to analyze time series in frequency domains.
By representing temporal dependencies as graph structures, these methods excel at capturing complex dependencies of temporal data. Despite their success in prediction tasks, their potential for generative tasks in synthetic time series generation has been largely overlooked. 

To address these limitations, we propose \textit{TSGDiff}, a novel framework for synthetic time series generation that rethinks the problem from a pure graph perspective. The core innovation of {\textit{TSGDiff}} lies in its unified approach, which combines dynamic graph construction and a diffusion-based generative process within the latent graph space. Nodes in the graph represent variables, and edges are constructed based on Fourier spectrum characteristics, enabling the model to capture intricate temporal dependencies and encode structural relationships in a flexible and adaptive manner. The diffusion-based process operates in the latent graph space, effectively modeling temporal semantic distributions and enabling the generation of time series with realistic dynamics and structural coherence.  
By unifying Fourier-based graph construction, latent representation learning, and diffusion-based modeling, {\textit{TSGDiff}} offers a robust and flexible solution for structured time series generation, ensuring high fidelity, temporal accuracy, and semantic coherence. Furthermore, we propose the Topological Structure Fidelity (Topo-FID) score, a novel graph-aware metric for assessing the similarity of time series graph structures. Topo-FID integrates two complementary sub-metrics: Graph Edit Similarity, which quantifies differences in adjacency matrices, and Structural Entropy Similarity, which evaluates the entropy of node degree distributions. This comprehensive metric bridges gaps in traditional evaluation methods, providing a more accurate and meaningful assessment of time series generation quality.

}

\lifeng{
To summarize, our main contributions include:  
\begin{itemize}  
    \item We propose \textit{TSGDiff}, the first framework to rethink synthetic time series generation from a graph-based perspective. By integrating a diffusion model within the latent graph space, it effectively captures and models the structural representation distribution of time series data.  
\end{itemize}  
\begin{itemize}  
    \item We introduce the Topological Structure Fidelity (Topo-FID), a graph-aware metric that quantifies the structural fidelity of the generated timeseries distribution.  
\end{itemize}  
\begin{itemize}  
    \item We validate the effectiveness of {\textit{TSGDiff}} on commonly-used real-world datasets, demonstrating its ability to generate high-quality time series data with realistic temporal and structural characteristics.  
\end{itemize}
}

\lifeng{
\section{Background}  
\subsection{Synthetic Time Series Generation}  

In recent years, a variety of methods based on different generative paradigms have emerged for time series generation \cite{naiman2024utilizing, crabbé2024time, zhou2023deep, park2024leveraging, alaa2021generative}. Generative Adversarial Network (GAN)-based approaches \cite{r:13a}, such as TimeGAN \cite{r:19x}, capture temporal dynamics by jointly optimizing supervised and adversarial objectives. Variational Autoencoder (VAE)-based methods \cite{r:14a} have been developed with specialized decoder structures tailored for time series data, incorporating trend and seasonal decomposition. 

With the rise of diffusion models in generative tasks, approaches like CSDI \cite{r:2a} and TimeDiff \cite{r:15a} have demonstrated remarkable performance in generating time series through iterative denoising processes. Specifically, CSDI introduces a self-supervised masking condition designed for imputation tasks, while TimeDiff develops an efficient future-mixup mechanism tailored for forecasting.  
For unconditional generation, Diffusion-TS mentioned in Section \ref{sec:intro} is one of the recent popular works. 
It models time series data distribution by a diffusion model enhanced by seasonal-trend decomposition and a frequency-enhanced objective. 
Despite their strengths, the above methods directly process raw time series data and fail to account for the temporal structured relationships from a pure graph perspective. 
}

\begin{figure*}[t!]
\centering
\includegraphics[width=0.94\textwidth]{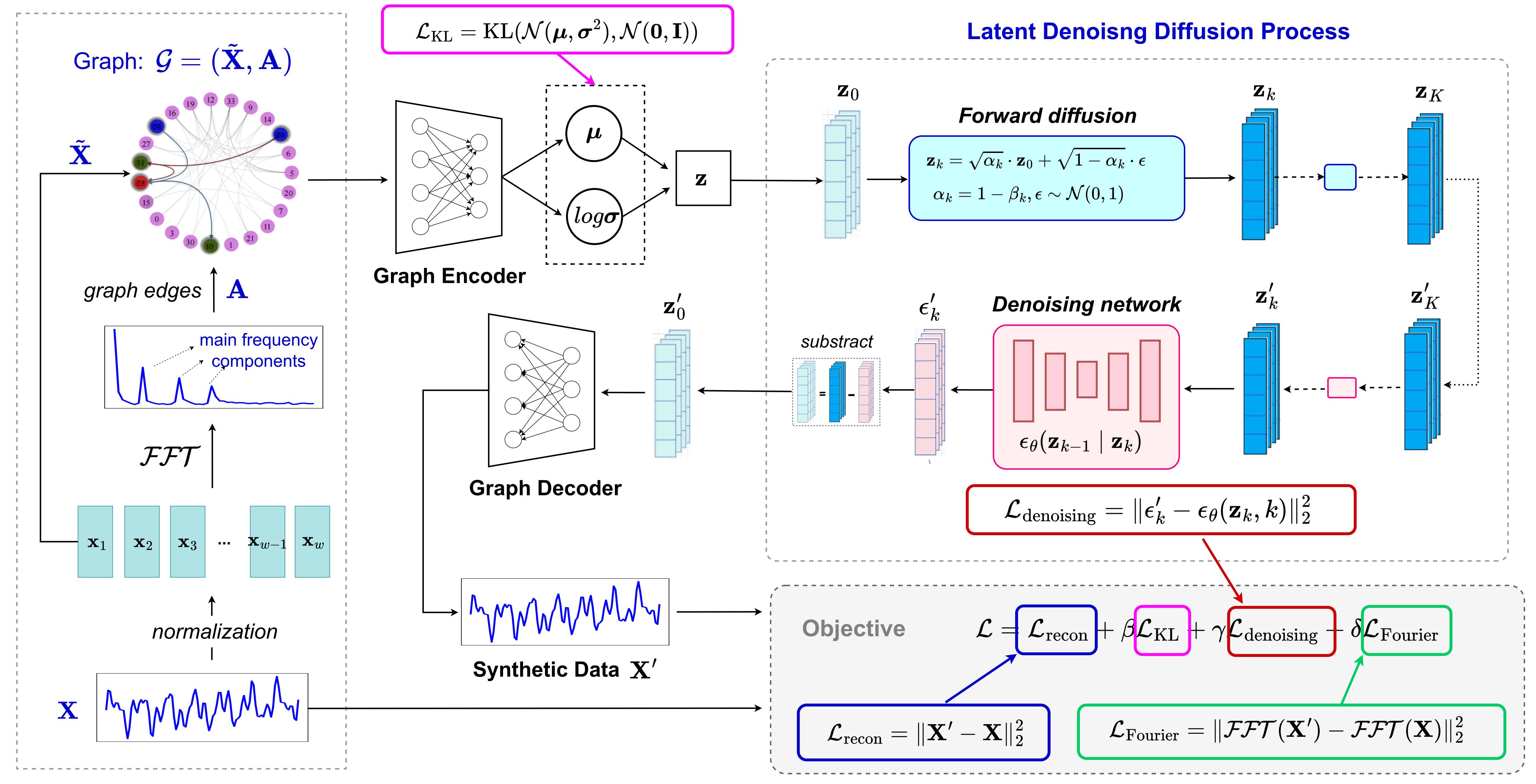}
\caption{Overview of the proposed \textit{TSGDiff} model.}
\label{fig:model}
\end{figure*}

\lifeng{
\subsection{Denoising Diffusion Probabilistic Models}
A diffusion probabilistic model \cite{r:9a} is designed to learn the reversal of a Markov chain process, known as the diffusion process, which incrementally adds noise to data, ultimately obliterating the original signal.

\noindent \textbf{Forward Diffusion.} Consider a data sample \( \rvx \in \mathbb{R}^d \sim p(\rvx) \) and a set of latent variables \(\{\rvx_0, \rvx_1, \cdots, \rvx_K\}\) ($\rvx_0=\rvx$) that interpolate between the data distribution and a Gaussian distribution as the diffusion steps progress. The forward process is formally defined as a Markov chain, parameterized by a sequence of variances \(\beta_k\) and \(\alpha_k := 1 - \beta_k\), and can be expressed as:
$q(\rvx_{1:K} \mid \rvx_0) = \prod_{k=1}^K q(\rvx_k \mid \rvx_{k-1})$,
where $q(\rvx_k \mid \rvx_{k-1}) = \mathcal{N}(\rvx_k; \sqrt{1-\beta_k}\rvx_{k-1}, \beta_k\mathbf{I})$.

As the number of diffusion steps increases, more noise is added to the data. Consequently, \( q(\rvx_k \mid \rvx) \) has a closed-form solution, which can be described in a general form:
\begin{equation}
\rvx_k = \sqrt{\overline{\alpha}_k} \rvx_0 + \sqrt{1-\overline{\alpha}_k}\,\rvepsilon, \quad \rvepsilon \sim \mathcal{N}(0, \mathbf{I}),
\end{equation}
where \(\overline{\alpha}_k = \prod_{i=1}^k (1 - \beta_i) \in (0, 1)\), \(\rvx_0 = \rvx\), and \(\rvx_K \sim \mathcal{N}(0, \mathbf{I})\). As the diffusion process progresses, the latent variable \(\rvx_k\) becomes increasingly noisy, eventually converging to \(\rvx_K\), which approximates a Gaussian distribution and becomes independent of the initial data sample \(\rvx\).

\noindent \textbf{Backward Denoising.} The denoising process is learnable and defined by the inverted Markov chain:  
\begin{equation}
p_{\theta}\left({\rvx}_{0:K}\right) = p({\rvx}_{K}) \prod_{k=1}^{K} p_{\theta}({\rvx}_{k-1}|{\rvx}_{k}),
\end{equation}
where \(p({\rvx}_{K})={\cal N}(0,\mathbf{I})\) is a known prior. The conditional distribution \(p_\theta (\rvx_{k-1}|\rvx_k)\) is approximated by a Gaussian distribution:  
\begin{equation}  
q(\rvx_{k-1}|\rvx_k,\rvx) = \mathcal{N}(\mu_k(\rvx_k,\rvx),\sigma_k^2\mathbf{I}),  
\end{equation}  
where \(\mu_k(\rvx_k,\rvx)\) has a closed-form solution and \(\sigma_k\) is a hyperparameter. Instead of predicting the clean data sample \(\rvx\) directly, the denoising network \(\epsilon_\theta\) is trained to estimate the noise \(\epsilon\) added during the forward process to the latent variable \(\rvx_k\). The noise-prediction loss function is:  
\begin{equation}  
L = \mathbb{E}_{\rvx \sim p(\rvx), k \sim \mathcal{U}\{1,\ldots,K\},\epsilon \sim \mathcal{N}(0, \mathbf{I})} \left[ \|\epsilon - \epsilon_\theta(\rvx_k, k)\|^2_2 \right],  
\end{equation}  
where \(\rvx_k = \sqrt{\alpha_k}\rvx + \sqrt{1-\alpha_k}\epsilon\), and \(\epsilon_\theta(\rvx_k, k)\) predicts the noise component \(\epsilon\).  

The conditional distribution \(p_\theta (\rvx_{k-1}|\rvx_k)\) is approximated as:
$p_\theta (\rvx_{k-1}|\rvx_k) \approx q(\rvx_{k-1} | \rvx_k, \hat{\rvx}_{\theta} (\rvx_k, k))$, 
where \(\hat{\rvx}_{\theta}(\rvx_k, k)\) is the implied clean sample estimate based on the predicted noise \(\epsilon_\theta(\rvx_k, k)\):  
\begin{equation}
\hat{\rvx}_{\theta}(\rvx_k, k) = \frac{\rvx_k - \sqrt{1-\alpha_k}\epsilon_\theta(\rvx_k, k)}{\sqrt{\alpha_k}}.
\end{equation}

\noindent \textbf{Sampling.} For sampling from the trained diffusion model, we follow the inference distribution proposed in \cite{song2020denoising}. The reverse process iteratively generates \(\rvx_{k-1}\) from \(\rvx_k\) using a Gaussian distribution parameterized by the mean and variance. The mean depends on both the current noisy variable \(\rvx_k\) and the predicted clean data \(\rvx\), while the variance controls stochasticity during sampling.   
When the variance is set to zero, the deterministic DDIM sampler is used, enabling efficient and stable sampling by directly denoising \(\rvx_k\) into a clean sample through a simplified closed-form update. This iterative process continues until the final reconstructed sample \(\rvx'_0\) is obtained. 

}

\section{\lifeng{Methodology}}
\lifeng{
This section formally elaborates on the proposed TSGDiff, a framework for synthetic time series generation that approaches the problem from a graph-based perspective. As shown in Figure \ref{fig:model}, the framework consists of three main stages: graph construction and encoding, latent diffusion-based modeling, and graph decoding. 
Initially, the time series data $\rmX$ is standardized and segmented using a sliding window approach. 
Each window $\tilde{\rmX}$ undergoes a Fourier transform to extract periodicity information, which is then used to construct a dynamic graph $\mathcal{G}(\tilde{\rmX},\rmA)$. In this graph, nodes represent variables, while edges in the adjacency matrix $\rmA$ are established based on characteristics of the Fourier spectrum, enabling the model to capture temporal dependencies and structural relationships effectively. 
The constructed graph is encoded into a latent representation using a graph neural network. Within this latent graph space, a diffusion-based generative process progressively refines noisy representations to recover meaningful temporal and structural patterns. This ensures the generated data maintains realistic dynamics and structural coherence. 
Finally, the denoised latent representation is decoded back into synthetic time series data, completing the generation process.

\subsection{Instance Normalization}

First, the sliding window technique divides long time series into fixed-length segments to extract local temporal patterns. Given a time series $\mathbf{X} = [\mathbf{x}_1, \mathbf{x}_2, \dots, \mathbf{x}_T]$, the segments are expressed as:  
$\rmX_{i:i+w} = [\mathbf{x}_i, \mathbf{x}_{i+1}, \dots, \mathbf{x}_{i+w-1}]$,
where $w$ denotes the window size
and each segment captures localized trends while maintaining temporal dependencies.  
To standardize features and enhance model training stability, Min-Max normalization is applied. This technique scales each feature of the time series to a target range, typically $[0, 1]$, by linearly transforming the data based on its minimum and maximum values, ensuring consistent feature scales across all samples.
For simplicity, we use $\tilde{\rmX}_{i}$ to represent the window $\rmX_{i:i+w}$.

}

\lifeng{
\subsection{Graph Construction}

To capture temporal dependencies \cite{r:23x}, each normalized segment $\tilde{\rmX}_{i}$ is transformed into a graph $\mathcal{G}$. In this graph, each time step in the slice corresponds to a node, and the node feature is the time series value at that time step.

Fourier transform\cite{r:86x} is utilized as the foundation for analyzing the frequency-domain characteristics of the time series. By decomposing the time-domain signal into its frequency components for each variable dimension independently, it enables the identification of periodicities that guide the graph construction process. For a time series slice $\tilde{\rmX}_{i}$, the discrete Fourier transform (DFT) for a specific variable is computed as:  
\begin{equation}
\tilde{X}_{i}^{(p)} = \sum_{n=0}^{w-1} \tilde{x}_{i+n} \cdot e^{-j\frac{2\pi}{w}pn}, \quad p = 0, 1, \dots, w-1,
\end{equation}  
where $\tilde{X}_{i}^{(p)}$ denotes the $p$-th frequency component, and $w$ represents the segment length (window size). The Fourier transform produces a frequency spectrum for each variable in the slice, highlighting the amplitude of its frequency components. The top three frequencies with the highest amplitudes are identified using a peak detection method and converted into their corresponding periods. These periods are then used to define edges in the graph, enabling the model to capture short-term, medium-term, and long-term variation patterns \cite{r:19x, s:17x, e:21x}.

The graph $\mathcal{G}$ is constructed by defining edges based on both temporal and periodic dependencies. To ensure the capture of local temporal relationships, an edge is always created between consecutive time steps. Additionally, edges are added based on the detected periods from the frequency spectrum, linking nodes that are periodic neighbors. This approach allows the graph to encode both short-term and long-term dependencies.  
The adjacency matrix $\mathbf{A}$ is used to represent the graph structure, where $\mathbf{A}_{ij} = 1$ indicates the presence of an edge between nodes $i$ and $j$, and $\mathbf{A}_{ij} = 0$ otherwise. The node feature matrix $\tilde{\mathbf{X}}$ stores the normalized time series values associated with each node. As shown in Figure \ref{fig:model} (left top), this graph construction process integrates temporal and periodic relationships into a cohesive structure, providing a rich representation of the time series to support downstream modeling tasks.
}

\lifeng{
\subsection{Graph Encoding}

The graph convolutional layer\cite{r:17x, r:70x} aggregates neighbor information by matrix multiplication between the adjacency matrix $\mathbf{A}$ and the node feature matrix $\tilde{\mathbf{X}}$. A single graph convolution operation is defined as:  
\begin{equation}
\mathbf{y}' = \text{Mish}(\text{BN}((W\mathbf{A}\tilde{\mathbf{X}} + b)^T)^T),
\end{equation}  
where $W$ and $b$ are learnable parameters, $\text{BN}(\cdot)$ represents batch normalization, and $\text{Mish}(\cdot)$ is the non-linear activation function. Residual connections are used to combine the input $\tilde{\mathbf{X}}$ with the output $\mathbf{y}'$, ensuring better gradient flow and preserving original features.

The graph encoder stacks multiple graph convolutional layers to process the input graph (node feature matrix $\tilde{\mathbf{X}}$ and adjacency matrix $\mathbf{A}$). After multiple layers of feature aggregation, mean pooling is applied to obtain the global graph representation $\mathbf{X}_{\text{pool}}$. The encoder then uses two linear layers to output the mean $\mathbf{\mu}$ and logarithmic standard deviation $\log\mathbf{\sigma}$ of the latent distribution:  
\begin{equation}
\mathbf{\mu} = \text{Linear}_\mu(\mathbf{X}_{\text{pool}}), \quad \log\mathbf{\sigma} = \text{Linear}_{\log\sigma}(\mathbf{X}_{\text{pool}}).
\end{equation}  

The latent vector $\mathbf{z}$ is sampled from the latent distribution using the reparameterization trick, ensuring differentiability:  
\begin{equation}
\mathbf{z} = \mathbf{\mu} + \mathbf{\epsilon} \odot \exp\left(\frac{\log\mathbf{\sigma}}{2}\right), \quad \mathbf{\epsilon} \sim \mathcal{N}(0, \mathbf{I}),
\end{equation}  
where $\mathbf{\epsilon}$ is sampled from a standard normal distribution, and $\odot$ denotes element-wise multiplication.

The graph decoder reconstructs the input graph from the latent vector $\mathbf{z}$. Starting with $\mathbf{z}$, fully connected layers with batch normalization and Mish activation progressively increase the feature dimension. The final output $\mathbf{\hat{X}}$ is computed using a linear transformation followed by the Tanh function to ensure the reconstructed graph matches the normalized range of the input:  
\begin{equation}
\mathbf{\hat{X}} = \text{Tanh}(\text{Linear}(\mathbf{z})).
\end{equation}  
This ensures the restoration of node features with dimensions consistent with the input graph data.

The graph encoder-decoder framework follows the principles of a Variational Autoencoder (VAE). A key component of this structure is the KL divergence loss $\mathcal{L}_{\text{KL}}$, which ensures that the latent variable distribution approximates a standard normal distribution $\mathcal{N}(0, \mathbf{I})$. The KL divergence loss is defined as:  
\begin{equation}
\mathcal{L}_{\text{KL}} = -\frac{1}{2} \sum_{j=1}^{d} \left( 1 + \log\sigma_j^2 - \mu_j^2 - \sigma_j^2 \right),
\end{equation}  
where $\mu_j$ and $\sigma_j^2$ are the mean and variance of the latent distribution for dimension $j$. 

This loss regularizes the latent space by encouraging the learned distribution $q_\phi(\mathbf{z}|\mathbf{X})$ to be close to the prior distribution $p(\mathbf{z}) = \mathcal{N}(0, \mathbf{I})$, ensuring smooth and meaningful representations in the latent space.
}

\lifeng{
\subsection{Latent Diffusion-based Modeling}

The KL divergence loss $\mathcal{L}_{\text{KL}}$ regularizes the latent variable $\mathbf{z}$ by constraining it to follow a standard normal distribution $\mathcal{N}(0, \mathbf{I})$. However, this constraint alone is insufficient for capturing the complex structure of data or enabling high-quality generative modeling. To address this, we introduce a latent diffusion-based approach, which refines and generates $\mathbf{z}$ through a denoising diffusion process.

\paragraph{Latent Diffusion Process.}

The forward process gradually corrupts $\mathbf{z}$ by adding Gaussian noise over time:
\begin{equation}
\mathbf{z}_k = \sqrt{\alpha_k} \cdot \mathbf{z} + \sqrt{1-\alpha_k} \cdot \mathbf{\epsilon}, \quad \mathbf{\epsilon} \sim \mathcal{N}(0, 1),
\end{equation}
where $\alpha_k = \prod_{s=1}^k \alpha_s$, $\alpha_s = 1 - \beta_s$, and $\beta_s$ is a noise coefficient. As $k$ increases, $\mathbf{z}_k$ approaches pure noise.

The reverse process recovers $\mathbf{z}$ by learning the posterior distribution $p_\theta(\mathbf{z}_{k-1} | \mathbf{z}_k, k)$:
\begin{equation}
p_\theta(\mathbf{z}_{k-1} | \mathbf{z}_k, k) := \mathcal{N}\left(\mathbf{z}_{k-1}; \mu_\theta(\mathbf{z}_k, k), \Sigma_\theta(k)\right),
\end{equation}
where $\mu_\theta$ is the predicted mean and $\Sigma_\theta(k)$ is the noise variance. The reverse update is:
\begin{equation}
\mathbf{z}_{k-1} = \sqrt{\frac{1}{\alpha_k}} \cdot \left( \mathbf{z}_k - \frac{1-\alpha_k}{\sqrt{1-\alpha_k}} \cdot (\mathbf{z}_k - \hat{\mathbf{z}}) \right) + \sqrt{\beta_k} \cdot \epsilon',
\end{equation}
where $\hat{\mathbf{z}} = \rvz_{\theta}(\mathbf{z}_k, k)$ is the predicted latent variable, and $\epsilon' \sim \mathcal{N}(0, 1)$ is reverse process noise.

\paragraph{Latent Diffusion Network.}

The reverse process is implemented using a latent diffusion network with an input layer, intermediate blocks, and an output layer:
\begin{equation}
\begin{aligned}
\mathbf{h}_0 &= {W}_{in}(\text{concat}(\mathbf{z}_k, \mathbf{t}_{\text{emb}})) + b_{\text{in}},  \\
\mathbf{h}_l &= \text{ReLU}\left({W}_l\mathbf{h}_{l-1} + b_l\right), \quad l = 1, \ldots, L,  \\
\hat{\mathbf{z}}_{k-1} &= {W}_{\text{out}}\mathbf{h}_L + b_{\text{out}}.
\end{aligned}
\end{equation}
Here, $\mathbf{z}_k$ is the noisy latent variable, $\mathbf{t}_{\text{emb}}$ is the time step embedding, and $\hat{\mathbf{z}}_{k-1}$ is the denoised prediction. 
Intuitively, Latent diffusion extends the KL-regularized latent space into a fully generative framework by iteratively denoising $\mathbf{z}$. This approach enables expressive modeling of complex data structures while supporting efficient and diverse sampling in the latent space.

}

\begin{table*}[t!]
  \centering
  \begin{tabular}{cccccccc} 
    \midrule[1pt]
    \textbf{Metric} & \textbf{Methods} & \texttt{ETTh} & \texttt{Stocks} & \texttt{Exchange} & \texttt{Weather} & \texttt{Wind} & \texttt{EEG} \\
    \midrule
    \multirow{5}{*}{\makecell[c]{Topo-FID $\nearrow$ \\ (\footnotesize{higher the better)}}} 
    & TSGDiff & \textbf{0.986$\pm$\textsubscript{0.0\%}} & \textbf{0.826$\pm$\textsubscript{0.0\%}} & \textbf{0.869$\pm$\textsubscript{0.0\%}} & \textbf{0.867$\pm$\textsubscript{0.0\%}} & \textbf{0.869$\pm$\textsubscript{0.0\%}} & \textbf{0.852$\pm$\textsubscript{0.0\%}} \\
    & Diffusion-TS & 0.864$\pm$\textsubscript{0.0\%} & 0.785$\pm$\textsubscript{0.0\%} & 0.806$\pm$\textsubscript{0.0\%} & 0.828$\pm$\textsubscript{0.0\%} & 0.819$\pm$\textsubscript{0.0\%} & 0.788$\pm$\textsubscript{0.0\%} \\
    & TimeGAN & 0.798$\pm$\textsubscript{0.0\%} & 0.795$\pm$\textsubscript{0.0\%} & 0.801$\pm$\textsubscript{0.0\%} & 0.816$\pm$\textsubscript{0.0\%} & 0.824$\pm$\textsubscript{0.0\%} & 0.828$\pm$\textsubscript{0.0\%} \\
    & Cot-GAN & 0.894$\pm$\textsubscript{0.0\%} & \underline{0.802$\pm$\textsubscript{0.0\%}} & \underline{0.809$\pm$\textsubscript{0.0\%}} & \underline{0.859$\pm$\textsubscript{0.0\%}} & 0.825$\pm$\textsubscript{0.0\%} & \underline{0.841$\pm$\textsubscript{0.0\%}} \\
    & TimeVAE & \underline{0.904$\pm$\textsubscript{0.0\%}} & 0.798$\pm$\textsubscript{0.0\%} & 0.804$\pm$\textsubscript{0.0\%} & 0.816$\pm$\textsubscript{0.0\%} & \underline{0.829$\pm$\textsubscript{0.0\%}} & 0.769$\pm$\textsubscript{0.0\%} \\
    \midrule
    \multirow{5}{*}{\makecell[c]{Context-FID $\searrow$ \\ \footnotesize (lower the better)}} 
    & TSGDiff & \textbf{0.224$\pm$\textsubscript{0.1\%}} & \textbf{0.357$\pm$\textsubscript{1.5\%}} & \textbf{0.061$\pm$\textsubscript{0.9\%}} & \textbf{0.353$\pm$\textsubscript{0.1\%}} & \textbf{0.256$\pm$\textsubscript{1.5\%}} & \textbf{0.020$\pm$\textsubscript{0.3\%}} \\
    & Diffusion-TS & \underline{0.225$\pm$\textsubscript{0.4\%}} & 0.530$\pm$\textsubscript{2.5\%} & 0.067$\pm$\textsubscript{1.0\%} & \underline{1.161$\pm$\textsubscript{0.1\%}} & \underline{0.491$\pm$\textsubscript{2.4\%}} & 0.030$\pm$\textsubscript{0.6\%} \\
    & TimeGAN & 1.392$\pm$\textsubscript{0.5\%} & \underline{0.377$\pm$\textsubscript{1.7\%}} & 1.103$\pm$\textsubscript{0.9\%} & 2.420$\pm$\textsubscript{2.3\%} & 5.087$\pm$\textsubscript{2.7\%} & \underline{0.023$\pm$\textsubscript{0.7\%}} \\
    & Cot-GAN & 3.486$\pm$\textsubscript{0.7\%} & 0.596$\pm$\textsubscript{0.8\%} & 1.523$\pm$\textsubscript{2.1\%} & 5.892$\pm$\textsubscript{1.2\%} & 5.409$\pm$\textsubscript{0.8\%} & 3.267$\pm$\textsubscript{0.5\%} \\
    & TimeVAE & 3.452$\pm$\textsubscript{0.6\%} & 0.830$\pm$\textsubscript{0.9\%} & \underline{0.065$\pm$\textsubscript{0.7\%}} & 13.952$\pm$\textsubscript{1.3\%} & 5.174$\pm$\textsubscript{0.0\%} & 8.874$\pm$\textsubscript{0.0\%} \\
    
    \midrule
    \multirow{5}{*}{\makecell[c]{Correlational $\searrow$ \\ \footnotesize (lower the better)}} 
    & TSGDiff & \textbf{0.024$\pm$\textsubscript{0.1\%}} & \textbf{0.026}$\pm$\textsubscript{0.1\%} & \textbf{0.019$\pm$\textsubscript{0.3\%}} & \textbf{0.035$\pm$\textsubscript{0.1\%}} & \textbf{0.022$\pm$\textsubscript{0.2\%}} & \textbf{0.201$\pm$\textsubscript{0.1\%}} \\
    & Diffusion-TS & \underline{0.024$\pm$\textsubscript{0.4\%}} & \underline{0.027$\pm$\textsubscript{0.1\%}} & 0.028$\pm$\textsubscript{0.5\%} & 0.054$\pm$\textsubscript{0.3\%} & \underline{0.033$\pm$\textsubscript{1.7\%}} & 0.466$\pm$\textsubscript{0.2\%} \\
    & TimeGAN & 0.126$\pm$\textsubscript{0.5\%} & 0.061$\pm$\textsubscript{0.7\%} & 0.131$\pm$\textsubscript{0.6\%} & 0.155$\pm$\textsubscript{0.1\%} & 0.454$\pm$\textsubscript{0.3\%} & 0.697$\pm$\textsubscript{0.2\%} \\
    
    & Cot-GAN & 0.032$\pm$\textsubscript{0.3\%} & 0.052$\pm$\textsubscript{1.3\%} & 0.031$\pm$\textsubscript{0.5\%} & 0.087$\pm$\textsubscript{0.9\%} & 0.043$\pm$\textsubscript{0.2\%} & 0.214$\pm$\textsubscript{0.1\%} \\
    & TimeVAE & 0.025$\pm$\textsubscript{0.2\%} & 0.053$\pm$\textsubscript{0.1\%} & \underline{0.022$\pm$\textsubscript{1.2\%}} & \underline{0.043$\pm$\textsubscript{0.7\%}} & 0.064$\pm$\textsubscript{0.5\%} & \underline{0.209$\pm$\textsubscript{1.2\%}} \\

    \midrule
    \multirow{5}{*}{\makecell[c]{Discriminative $\searrow$ \\ \footnotesize (lower the better)}} 
    & TSGDiff & \textbf{0.056$\pm$\textsubscript{0.6\%}} & \textbf{0.019$\pm$\textsubscript{1.3\%}} & \textbf{0.052$\pm$\textsubscript{0.9\%}} & \textbf{0.283$\pm$\textsubscript{0.1\%}} & \textbf{0.073$\pm$\textsubscript{0.3\%}} & \textbf{0.301$\pm$\textsubscript{1.5\%}} \\
    & Diffusion-TS & \underline{0.111$\pm$\textsubscript{0.1\%}} & \underline{0.075$\pm$\textsubscript{0.5\%}} & 0.172$\pm$\textsubscript{0.9\%} & 0.398$\pm$\textsubscript{1.8\%} & \underline{0.161$\pm$\textsubscript{1.1\%}} & 0.492$\pm$\textsubscript{1.3\%} \\
    & TimeGAN & 0.353$\pm$\textsubscript{0.7\%} & 0.193$\pm$\textsubscript{2.5\%} & 0.465$\pm$\textsubscript{1.5\%} & 0.494$\pm$\textsubscript{1.8\%} & 0.495$\pm$\textsubscript{0.5\%} & 0.391$\pm$\textsubscript{1.7\%} \\
    & Cot-GAN & 0.114$\pm$\textsubscript{1.5\%} & 0.102$\pm$\textsubscript{1.6\%} & 0.154$\pm$\textsubscript{2.5\%} & \underline{0.297$\pm$\textsubscript{0.8\%}} & 0.168$\pm$\textsubscript{0.7\%} & \underline{0.335$\pm$\textsubscript{1.1\%}} \\
    & TimeVAE & 0.346$\pm$\textsubscript{1.6\%} & 0.260$\pm$\textsubscript{3.3\%} & \underline{0.125$\pm$\textsubscript{2.1\%}} & 0.496$\pm$\textsubscript{1.9\%} & 0.338$\pm$\textsubscript{0.9\%} & 0.496$\pm$\textsubscript{2.7\%} \\
    
    \midrule
    \multirow{5}{*}{\makecell[c]{Predictive $\searrow$ \\ \footnotesize (lower the better)}} 
    & TSGDiff & \textbf{0.020$\pm$\textsubscript{0.1\%}} & \textbf{0.005$\pm$\textsubscript{0.1\%}} & \textbf{0.004$\pm$\textsubscript{0.0\%}} & \textbf{0.009$\pm$\textsubscript{0.0\%}} & \textbf{0.007$\pm$\textsubscript{0.0\%}} & \textbf{0.001$\pm$\textsubscript{0.1\%}} \\
    & Diffusion-TS & \underline{0.026$\pm$\textsubscript{0.0\%}} & \underline{0.005$\pm$\textsubscript{0.2\%}} & 0.005$\pm$\textsubscript{0.2\%} & 0.011$\pm$\textsubscript{0.0\%} & \underline{0.009$\pm$\textsubscript{0.0\%}} &  \underline{0.002$\pm$\textsubscript{0.2\%}} \\
    & TimeGAN & 0.030$\pm$\textsubscript{0.2\%} & 0.006$\pm$\textsubscript{0.1\%} & \underline{0.004$\pm$\textsubscript{0.1\%}} & \underline{0.009$\pm$\textsubscript{0.1\%}} & 0.009$\pm$\textsubscript{0.1\%} & 0.002$\pm$\textsubscript{0.3\%} \\
    & Cot-GAN & 0.028$\pm$\textsubscript{0.2\%} & 0.013$\pm$\textsubscript{0.1\%} & 0.006$\pm$\textsubscript{0.3\%} & 0.021$\pm$\textsubscript{0.2\%} & 0.026$\pm$\textsubscript{0.2\%} & 0.035$\pm$\textsubscript{0.3\%} \\
    & TimeVAE & 0.038$\pm$\textsubscript{0.2\%} & 0.016$\pm$\textsubscript{0.2\%} & 0.006$\pm$\textsubscript{0.1\%} & 0.031$\pm$\textsubscript{0.3\%} & 0.022$\pm$\textsubscript{0.0\%} & 0.039$\pm$\textsubscript{0.2\%} \\
    \bottomrule
  \end{tabular}
  \caption{\lifeng{Results of multivariate time-series generation (bold indicates best performance, while underline is the second best).}}
  \label{tab:time_series_results}
\end{table*}

\lifeng{
\subsubsection{Training Objective}

The denoising loss measures the difference between the noisy latent variable and the output of the denoising network $\mathbf{\epsilon}_\theta$, which predicts clean latent variables:
\begin{equation}
L_{\text{denoising}} = \mathbb{E}_{k, \mathbf{z}_k, \mathbf{\epsilon}} \left[ \| \mathbf{\epsilon} - \mathbf{\epsilon}_\theta(\mathbf{z}_k, k) \|^2 \right],
\end{equation}
where $\mathbf{\epsilon}_\theta$ is the noise estimated by the denoising network, $\mathbf{z}_k$ is the noisy latent variable at time step $k$, and $\mathbf{\epsilon}$ is the actual noise added during the forward process.

To further enhance consistency, we incorporate a Fourier loss inspired by Diffusion-TS, which enforces alignment in the frequency domain:
\begin{equation}
L_{\text{Fourier}} = \mathbb{E}_{\mathbf{x}_0} \| \mathcal{FFT}(\mathbf{x}_0) - \mathcal{FFT}(\hat{\mathbf{x}}_0) \|^2,
\end{equation}
where $\mathcal{FFT}$ represents the Fourier transform, $\mathbf{x}_0$ is the original data, and $\hat{\mathbf{x}}_0$ is the reconstruction.

The total loss integrates all components:
\begin{equation}
\mathcal{L} = \mathcal{L}_{\text{recon}} + \beta \mathcal{L}_{\text{KL}} + \gamma \mathcal{L}_{\text{denoising}} + \delta \mathcal{L}_{\text{Fourier}},
\end{equation}
where $\mathcal{L}_{\text{recon}}$ is the reconstruction loss capturing the MSE between the input data and decoder output,
and $\beta, \gamma, \delta$ are weights balancing each term.
In practice, we set \(\beta = 0.2\), while \(\gamma = 1\) and \(\delta = 1\).
By minimizing $\mathcal{L}$, the model learns to denoise latent variables, enforce frequency consistency, and reconstruct time series data accurately, ensuring the generated data aligns with the original distribution.
}

\section{Experiments}
\lifeng{
\subsection{Setup}
\paragraph{Datasets.} 
We use six public real-world time series datasets from diverse domains:
i) \texttt{ETTh}: Hourly energy and environmental data (2016–2018) with over 15,000 time steps.  
ii) \texttt{Stocks}\footnote{\url{https://finance.yahoo.com/quote/GOOG}}: Daily multi-stock data with prices and trading volumes spanning multiple years.  
iii) \texttt{Exchange}\footnote{\url{https://github.com/laiguokun/multivariate-time-series-data}}: Daily USD exchange rates (1990–2010) against major currencies.  
iv) \texttt{Weather}\footnote{\url{https://www.bgc-jena.mpg.de/wetter/}}: Meteorological data recorded every 10 minutes, with over 50,000 time steps and 20+ features like temperature and humidity.  
v) \texttt{Wind}\footnote{ https://github.com/PaddlePaddle/PaddleSpatial/tree/main/ paddlespatial/datasets/WindPower}: Hourly wind farm generation data across seasons, spanning over a year.  
vi) \texttt{EEG}\footnote{\url{https://archive.ics.uci.edu/dataset/264/eeg+eye+state}}: High-frequency (128Hz) \texttt{EEG} signals with over 15,000 entries. 
}

\begin{figure*}[htbp]
    \centering
    \setlength{\tabcolsep}{0pt}  
    
    \begin{minipage}[b]{0.24\textwidth}  
        \centering
        \includegraphics[width=\textwidth]{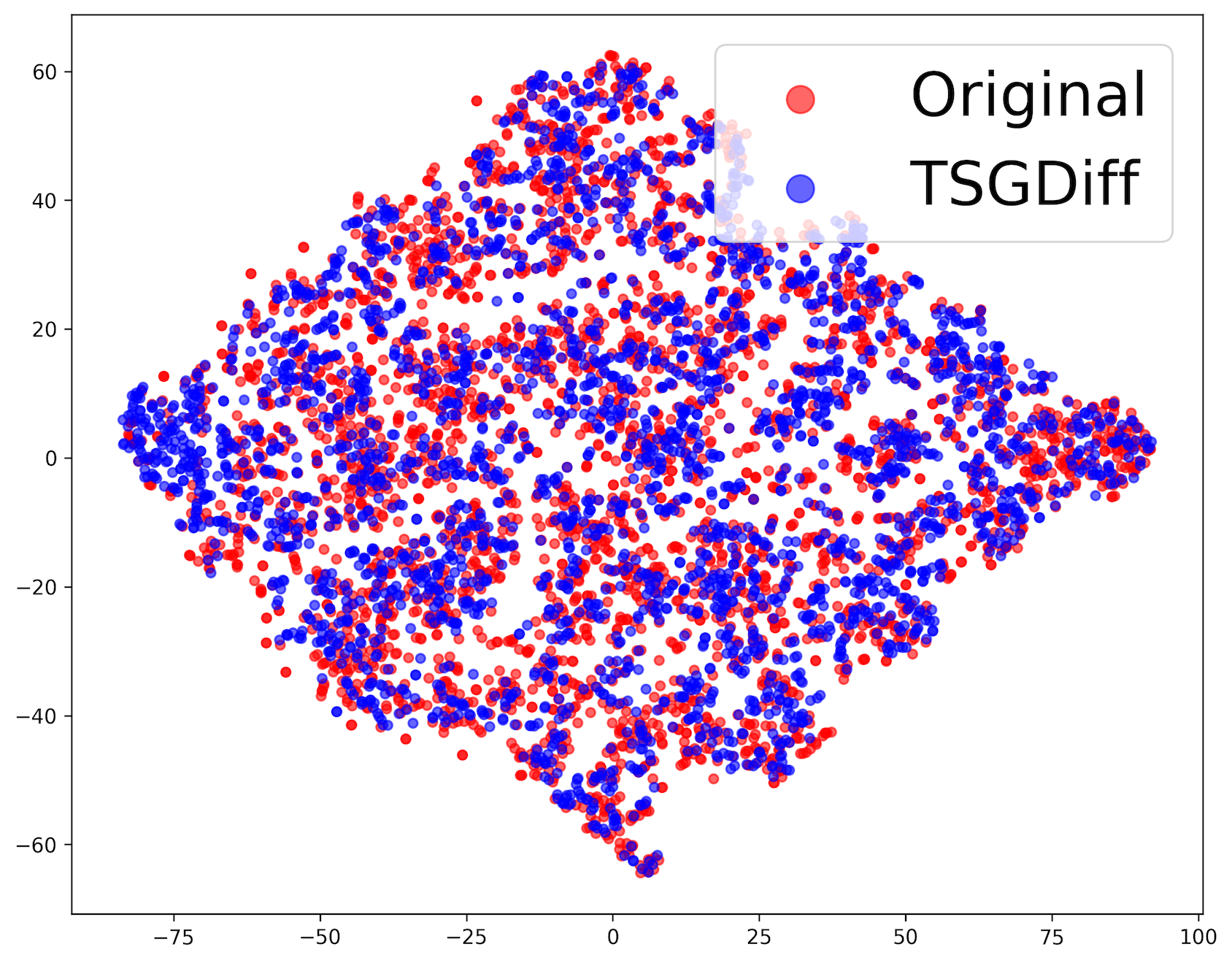}\\  
        \includegraphics[width=0.95\textwidth]{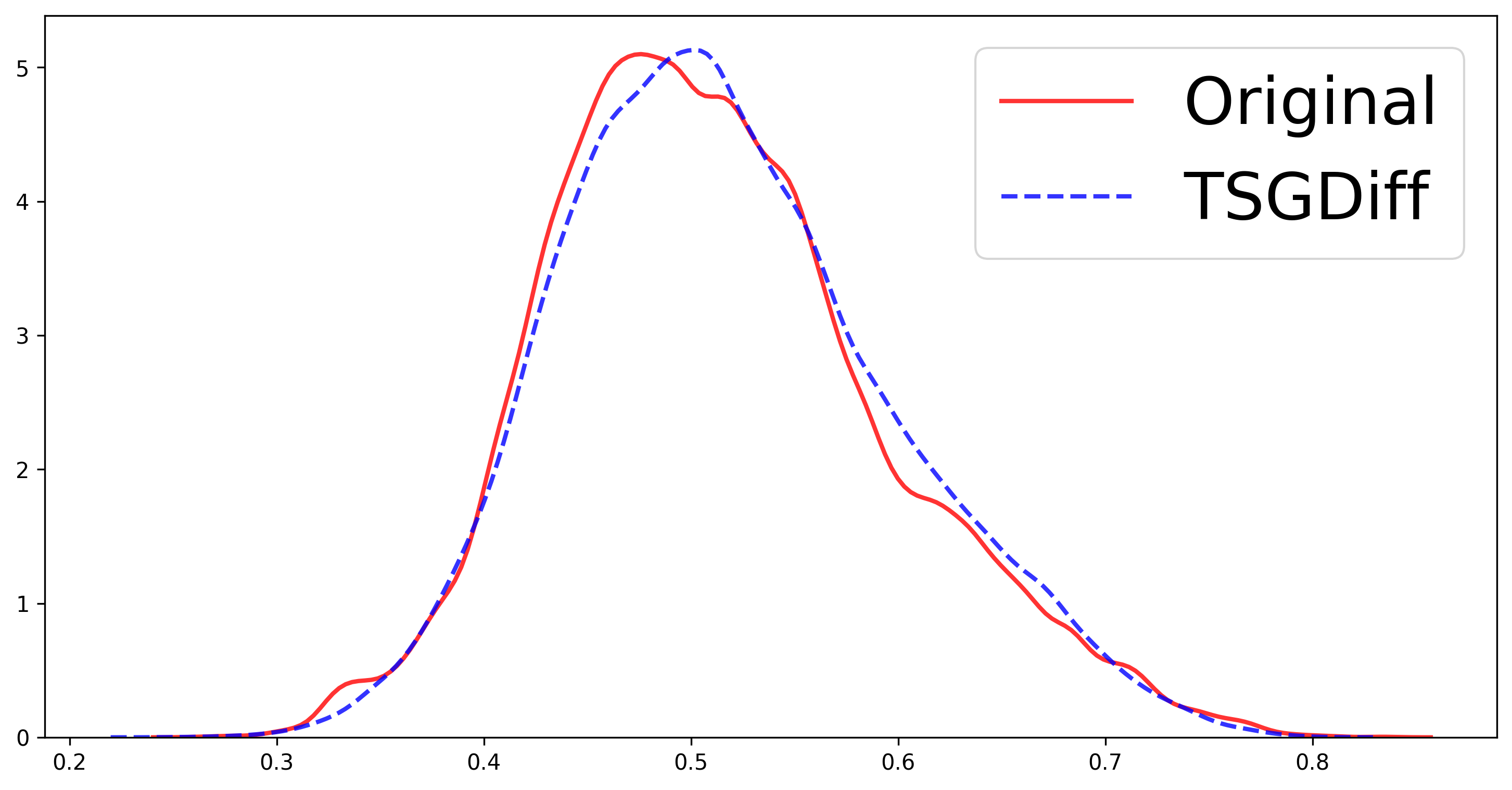}
        \caption*{(a) Ours on \texttt{ETTh}}
    \end{minipage}%
    \hfill  
    \begin{minipage}[b]{0.24\textwidth}
        \centering
        \includegraphics[width=\textwidth]{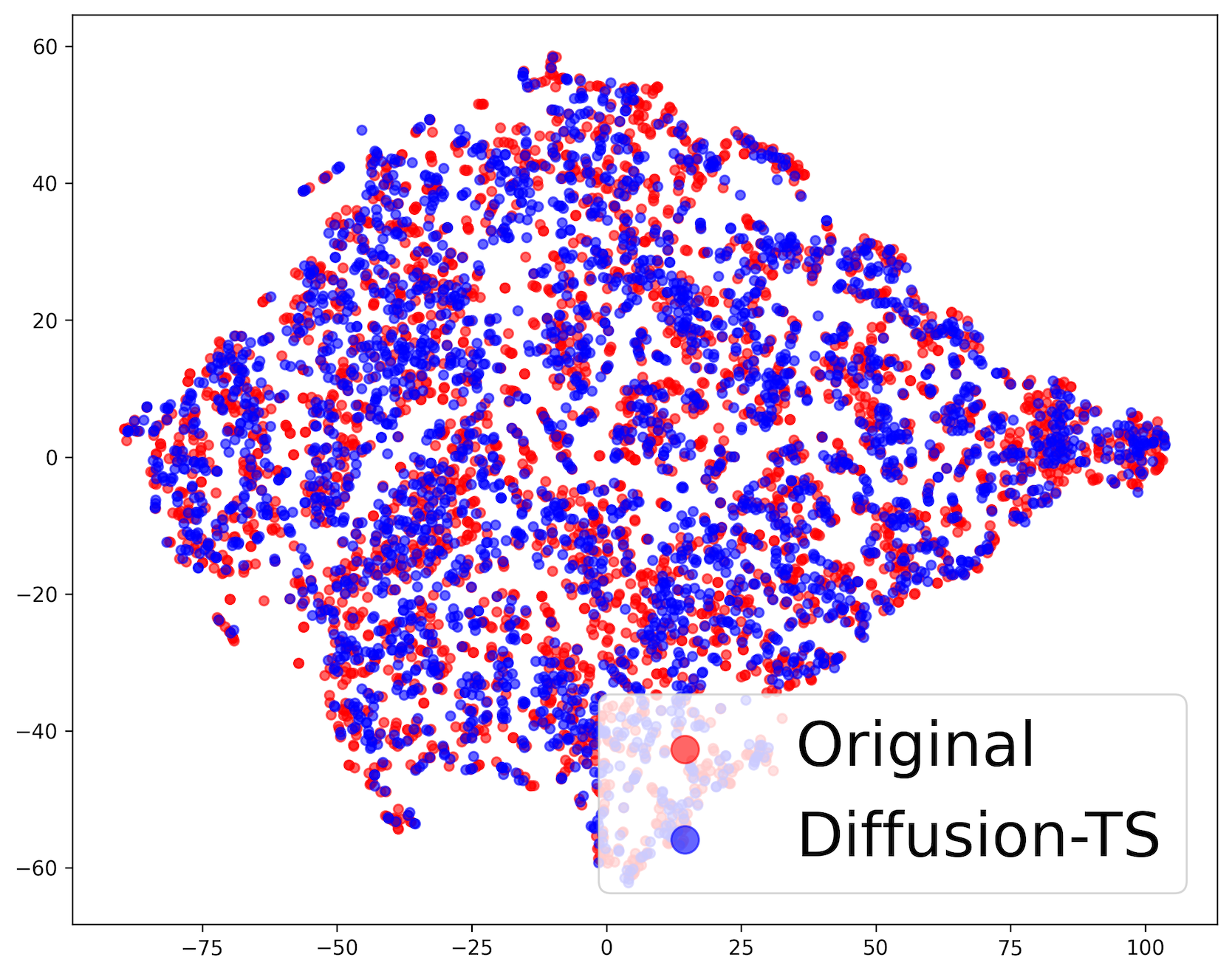}  
        \includegraphics[width=0.96\textwidth]{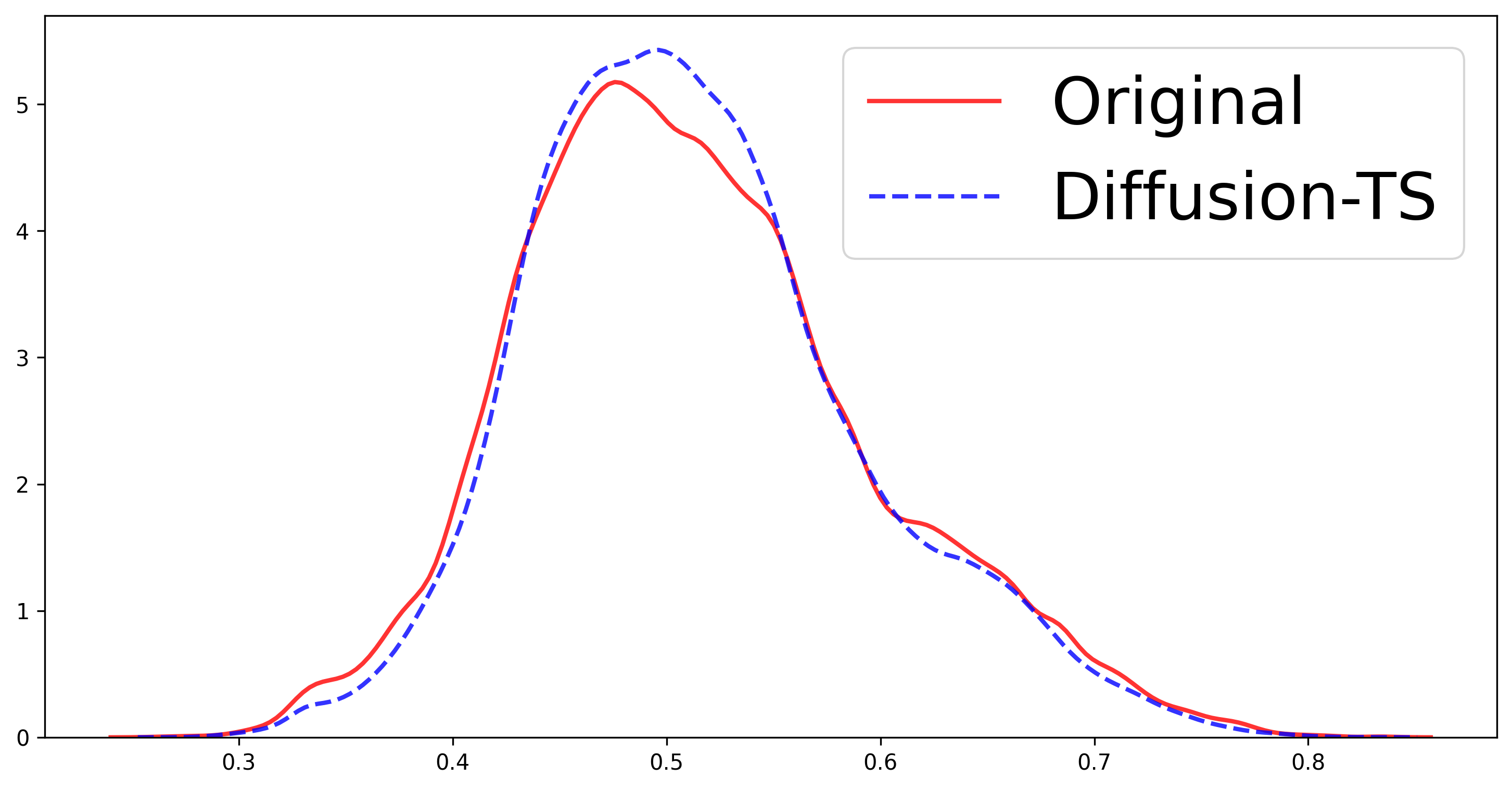}
        \caption*{(b) Diffusion-TS on \texttt{ETTh}}
    \end{minipage}%
    \hfill
    \begin{minipage}[b]{0.24\textwidth}
        \centering
        \includegraphics[width=\textwidth]{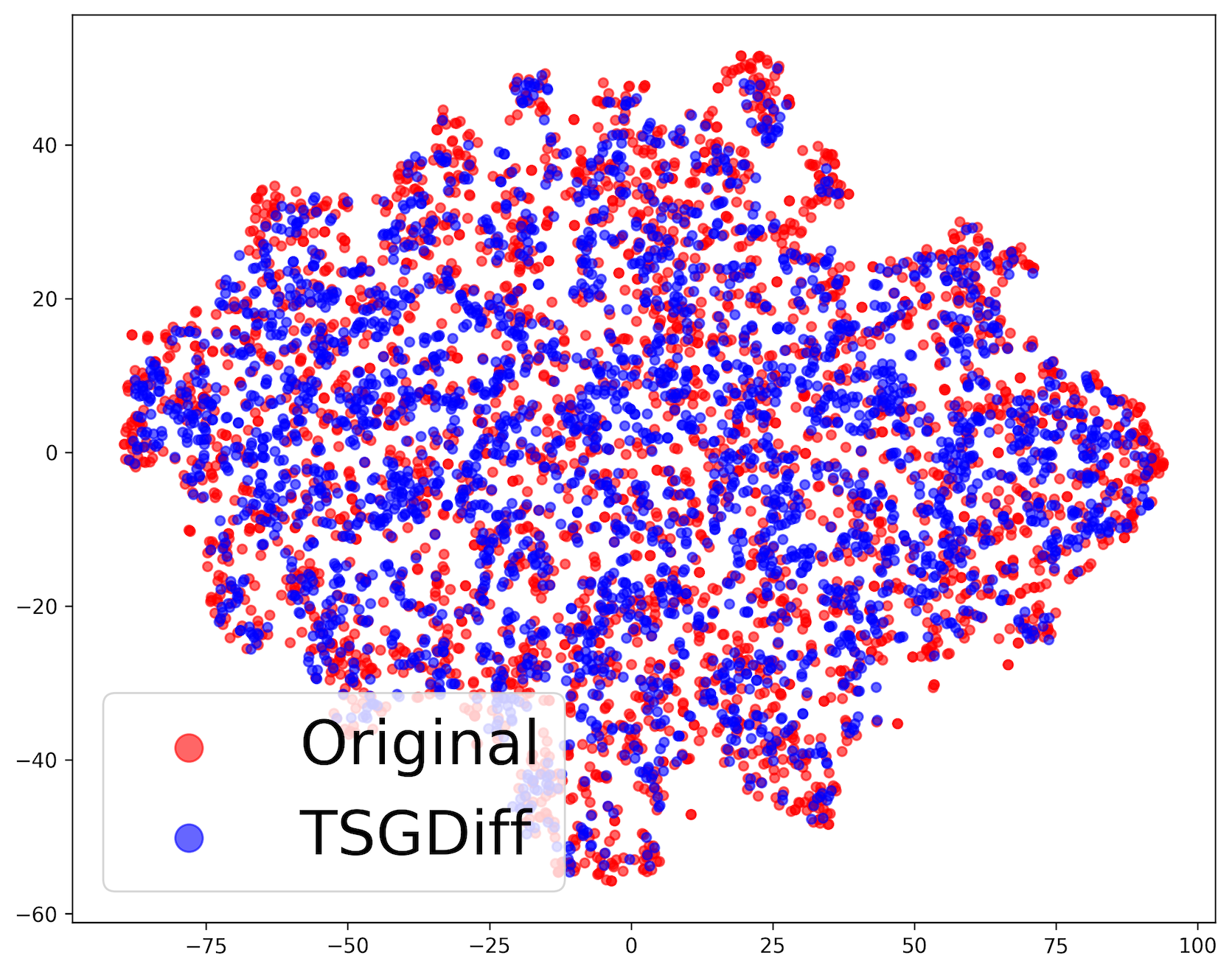}  
        \includegraphics[width=0.96\textwidth]{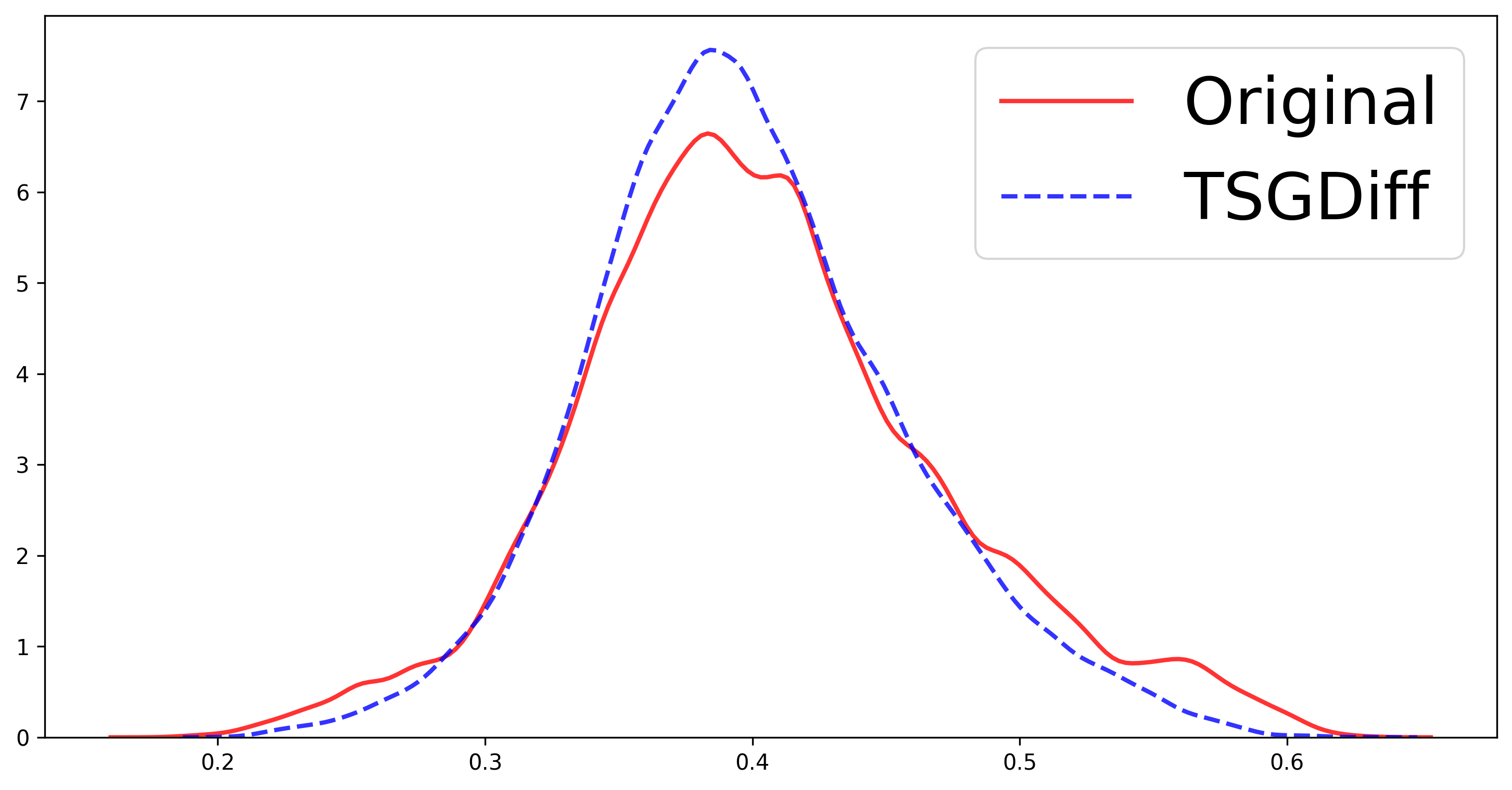}
        \caption*{(c) Ours on \texttt{Wind}}
    \end{minipage}%
    \hfill
    \begin{minipage}[b]{0.24\textwidth}
        \centering
        \includegraphics[width=\textwidth]{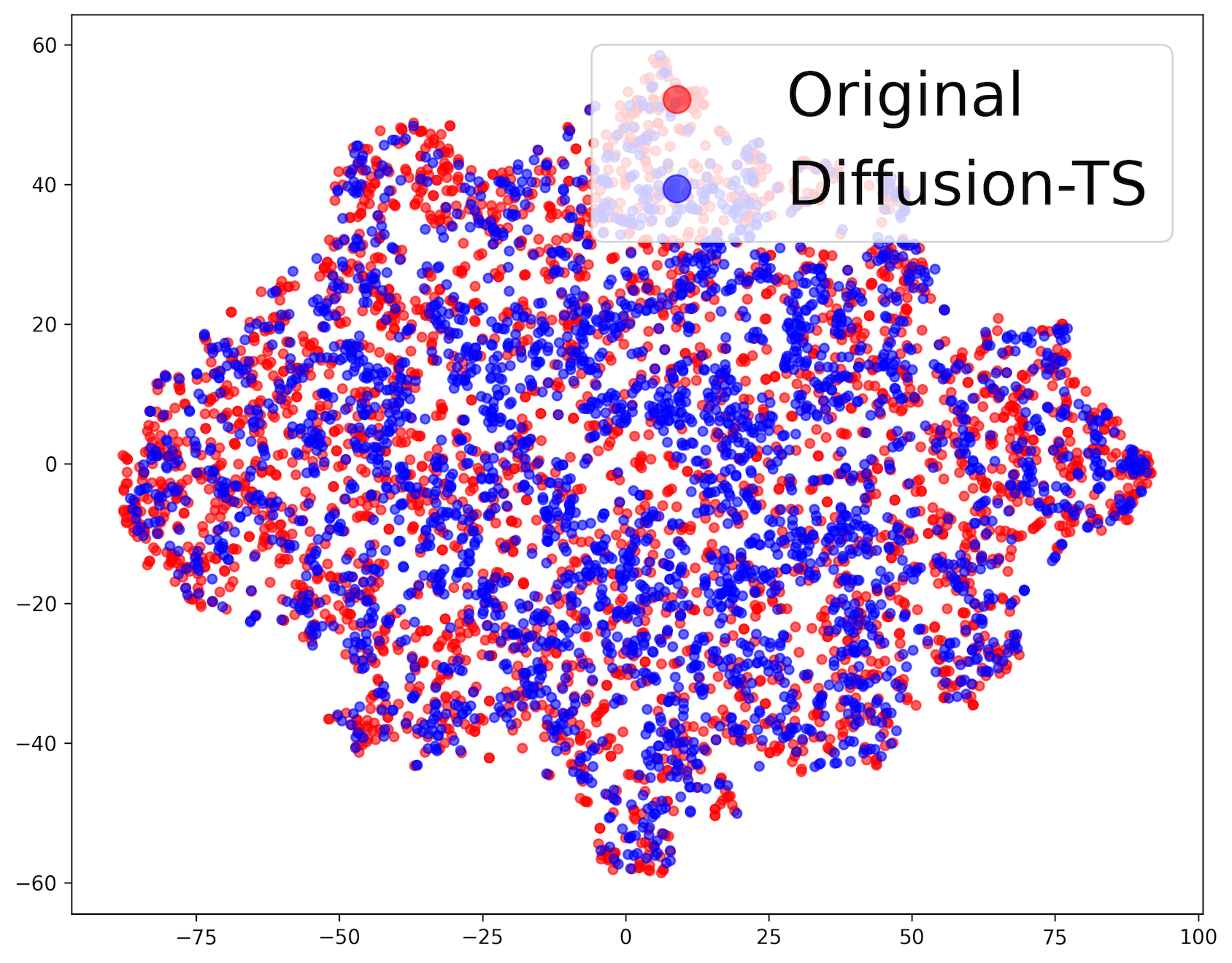}  
        \includegraphics[width=0.95\textwidth]{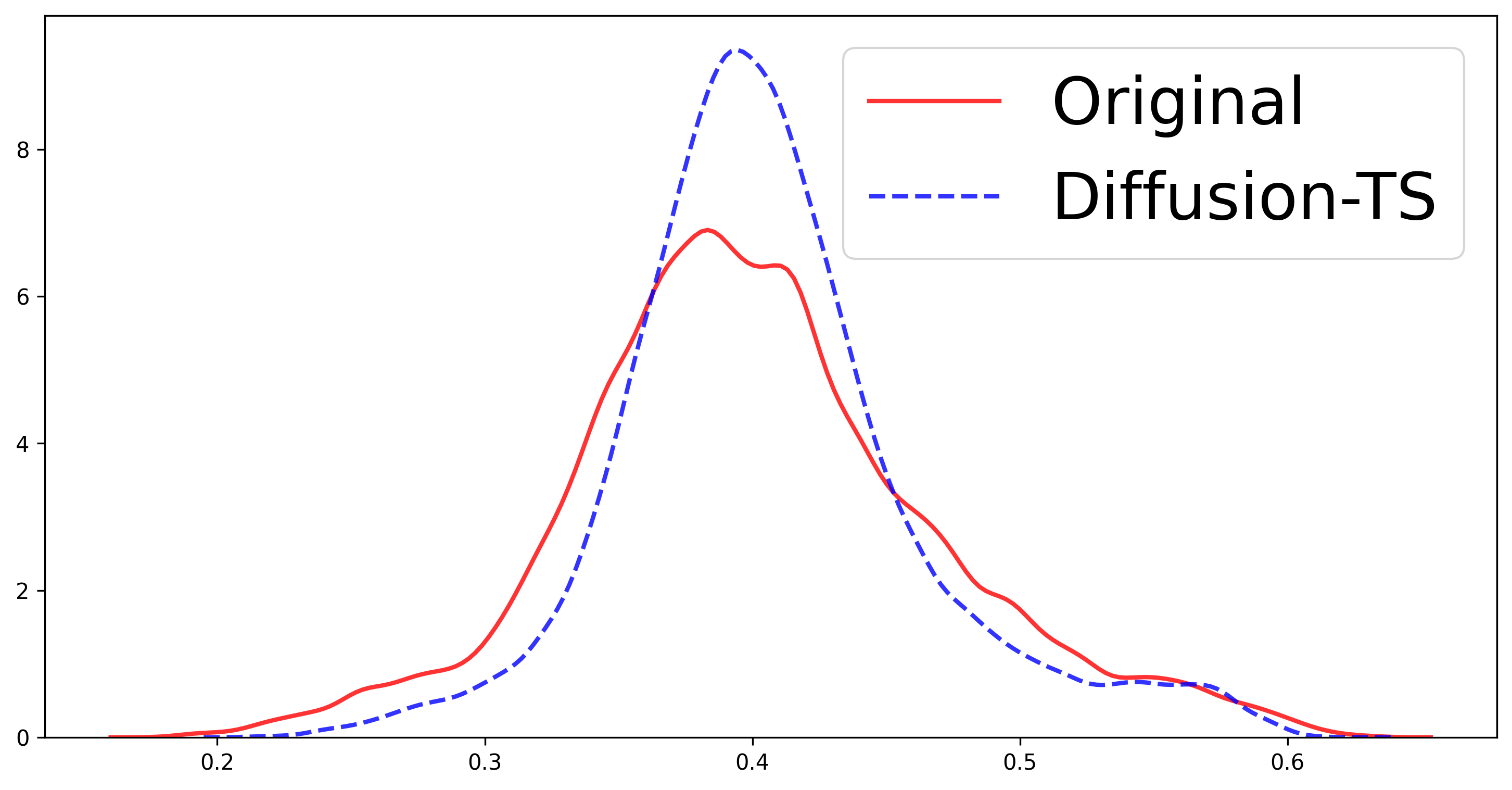}
        \caption*{(d) Diffusion-TS on \texttt{Wind}}
    \end{minipage}
    
    \caption{Visualizations of the time series synthesized by \textit{TSGDiff} and Diffusion-TS.}
    \label{fig:time_series_visualization}
\end{figure*}

\lifeng{
\paragraph{Evaluation Metrics.} 
We evaluate synthetic data using four traditional metrics: Context-FID \cite{r:22x} (local context quality), Correlational score \cite{r:20x} (temporal dependency via cross-correlation), Discriminative score \cite{r:19x} (classifier-based similarity), and Predictive score \cite{r:19x} (TSTR-based utility).

These metrics, however, fail to capture the intrinsic structural and frequency characteristics of time series data. To address this, we propose Topological Structure Fidelity (Topo-FID), a new metric that assesses the similarity of distributions between time series samples by evaluating their graph representations \(\mathcal{G}\). 
Formally, Topo-FID is defined as the expected similarity of graph representations derived from pairs of sampled time series:
\begin{equation}
\text{Topo-FID} = \mathbb{E}_{(\mathcal{G}, \hat{\mathcal{G}}) \sim P} \big[ \alpha \cdot S_\text{edit} + (1-\alpha) \cdot S_\text{entropy} \big],
\end{equation}
where \(\mathcal{G}\) and \(\hat{\mathcal{G}}\) are the graph representations of real and synthetic time series samples, \(P\) is the sampling distribution. $\alpha$ is fixed to be 0.5 by default. 
Here, \(S_\text{edit}\) is the Graph Edit Similarity, which measures edge structure similarity:
\begin{equation}
S_\text{edit} = 1 - \frac{\sum_{i,j} |\mathbf{A}_{i,j} - \hat{\mathbf{A}}_{i,j}|}{N^2},
\end{equation}
where \(\mathbf{A}\) and \(\hat{\mathbf{A}}\) are adjacency matrices, and \(N\) is the number of nodes. 
\(S_\text{entropy}\) is the Structural Entropy Similarity, which evaluates node degree distribution similarity:
\begin{equation}
S_\text{entropy} = \frac{1}{1 + |H(\mathbf{A}) - H(\hat{\mathbf{A}})|},
\end{equation}
with \(H(\mathbf{A})=-\sum_{d \in D} p(d) \log_2(p(d) + \epsilon)\),  
where \(p(d)\) is the degree distribution of $\mathbf{A}$, and \(\epsilon = 10^{-10}\) avoids logarithmic errors.
}

\lifeng{
\paragraph{Baselines.}
In experiments, we have selected four popular and strong baselines for comparison, including:  
i) a diffusion model, Diffusion-TS \cite{r:24x};  
ii) two GAN-based models, TimeGAN \cite{r:19x} and Cot-GAN \cite{r:20x};  
iii) a variational autoencoder, TimeVAE \cite{c:23}. 

\paragraph{Implementation Details.}
In the Graph Encoder, we use three stacked GraphConvBlocks, with both the hidden layer and latent space dimensions set to 1600. The Diffusion model employs three DiffusionBlocks with 64 units each (where each DiffusionBlock is a three-layer MLP network), and the timesteps are set to 1000. The Graph Decoder uses four fully connected layers to restore the 1600-dimensional features to their original dimension. 
The sliding window size is 48 (say, 2 days for \texttt{ETTh}, 48 days for \texttt{Stocks}, 48 days for \texttt{Exchange}, 8h for \texttt{Weather}, 6h for \texttt{Wind}, 375ms for \texttt{EEG}), with a stride of 1.   
We use a batch size of 128, a learning rate of 0.01, and train for 500 epochs. 
All experiments are repeated five times, implemented in PyTorch \cite{paszke2019pytorch}, and conducted on an NVIDIA RTX 4090 GPU with 24GB memory.
}

\begin{table*}[t!]
  \centering
  \begin{tabular}{>{\centering\arraybackslash}cccccc} 
    \toprule
    \textbf{Method} & \textbf{Topo-FID$\uparrow$} & \textbf{Context-FID$\downarrow$} & \textbf{Correlational$\downarrow$} & \textbf{Discriminative$\downarrow$} & \textbf{Predictive$\downarrow$} \\
    \midrule
    TSGDiff
    & \textbf{0.986$\pm$\textsubscript{0.0\%}} & \textbf{0.224$\pm$\textsubscript{0.1\%}} & \textbf{0.024$\pm$\textsubscript{0.1\%}} & \textbf{0.056$\pm$\textsubscript{0.6\%}} & \textbf{0.020$\pm$\textsubscript{0.1\%}} \\
    
    \midrule
    w/o KL
    & 0.787$\pm$\textsubscript{0.0\%} & 44.467$\pm$\textsubscript{1.5\%} & 0.159$\pm$\textsubscript{0.1\%} & 0.500$\pm$\textsubscript{1.3\%} & 0.064$\pm$\textsubscript{0.1\%} \\
    
    \midrule
    w/o Denoising
    & 0.908$\pm$\textsubscript{0.0\%} & 3.922$\pm$\textsubscript{0.9\%} & 0.146$\pm$\textsubscript{0.3\%} & 0.432$\pm$\textsubscript{0.9\%} & 0.019$\pm$\textsubscript{0.2\%} \\

    \midrule
    w/o Fourier
    & 0.883$\pm$\textsubscript{0.0\%} & 0.407$\pm$\textsubscript{0.1\%} & 0.024$\pm$\textsubscript{0.1\%} & 0.061$\pm$\textsubscript{0.7\%} & 0.021$\pm$\textsubscript{0.3\%} \\
    \bottomrule
  \end{tabular}
  \caption{Ablation study results on \texttt{ETTh} dataset (Bold indicates the best).}
  \label{tab:ABLATION}
\end{table*}



\lifeng{
\subsection{Results for Synthetic Time Series Generation}
Table \ref{tab:time_series_results} highlights the performance of \textit{TSGDiff} compared to baselines, including Diffusion-TS \cite{r:24x}, TimeGAN \cite{r:19x}, Cot-GAN \cite{r:20x}, and TimeVAE \cite{c:23}. The results show that \textit{TSGDiff} consistently produces higher-quality synthetic data across nearly all metrics, improving the average Discriminative score by 50\% across six datasets.

Compared to Diffusion-TS, \textit{TSGDiff} leverages graph-based latent space modeling to capture structural temporal dependencies better, excelling on complex datasets like \texttt{Weather} and \texttt{EEG}. Against GAN-based models such as TimeGAN and Cot-GAN, it avoids mode collapse, producing more robust and diverse samples. Relative to TimeVAE, \textit{TSGDiff} preserves structural coherence more effectively through its unified graph-based framework.

These advantages are especially evident in the Topo-FID metric, where \textit{TSGDiff} significantly outperforms all baselines, demonstrating its ability to generate data with superior structural fidelity and temporal accuracy.

}

\lifeng{
\subsection{Visualization Results}

We used two visualization methods to evaluate time series synthesis performance as suggested by \cite{r:24x}:  
i) projecting original and synthetic data into a two-dimensional space using t-SNE \cite{r:08x}, and  
ii) visualizing data distributions via kernel density estimation.   
As shown in Figure \ref{fig:time_series_visualization}, while the first row (t-SNE) does not clearly indicate which method performs better, the second row (kernel density estimation) demonstrates that the synthetic distribution generated by \textit{TSGDiff} closely matches the original data, outperforming Diffusion-TS.
}

\lifeng{
\subsection{Ablation Study}

To assess the contributions of different components in \textit{TSGDiff}, we compared the full model with three variants:  
i) \textit{w/o KL divergence loss}: Removes the KL divergence loss to analyze its role in constraining latent variable distributions.  
ii) \textit{w/o Denoising loss}: Excludes the diffusion model loss, evaluating its contribution to generative capability.  
iii) \textit{w/o Fourier loss}: Omits the Fourier loss term to examine its impact on capturing frequency-domain features.

As shown in Table \ref{tab:ABLATION}, the KL divergence loss, denoising loss, and Fourier loss each play a critical role in \textit{TSGDiff}'s performance. The KL divergence loss provides a foundational prior by aligning latent variables with a standard normal distribution, ensuring diversity and stability while preventing mode collapse and unstable training. Building on this foundation, the denoising loss further refines latent variables during the reverse diffusion process, preserving the fidelity of generated samples. Meanwhile, the Fourier loss enhances reconstruction precision by enforcing frequency-domain consistency, preserving key spectral features like periodicity. Together, these components form a cohesive framework that ensures robust performance, high-quality data reconstruction, and effective handling of complex time series synthesis tasks.

}

\lifeng{
\subsection{Efficiency Analysis}

We analyzed \textit{TSGDiff}'s graph construction time cost 
to evaluate the efficiency of transforming time series into graphs.

As shown in Table \ref{tab:Efficiency}, the graph construction process is highly efficient, with time costs consistently under 5 seconds and accounting for less than 0.1\% of the total training time. This demonstrates that the process introduces negligible overhead, ensuring smooth integration into the overall pipeline. 
This efficiency aligns seamlessly with the latent diffusion framework in \textit{TSGDiff}, which operates in a compressed latent space to balance computational cost and model performance. The lightweight yet expressive graph representations enable the latent diffusion process to focus on capturing complex temporal dependencies, while maintaining high-quality reconstructions. Together, these components highlight the practicality and scalability of \textit{TSGDiff} in handling large-scale datasets and complex time series tasks.

\begin{table}[t!]
  \centering
  \footnotesize
  \begin{tabular}{ccccc} 
    \midrule[1pt]
     & \texttt{ETTh} & \texttt{Stocks} & \texttt{Weather} \\
    \midrule
    \# iterations & 68,000 & 14,500 & 61,000 \\
    \midrule
    graph construction (s)
    & 3.12 & 0.82  & 3.79 \\
    total training (s)
    & 4,506.15 & 970.23  & 4025.31 \\\midrule
    ratio (\%) & 0.07 & 0.08  & 0.09 \\
    \bottomrule
  \end{tabular}
  \caption{Efficiency analysis. Here, ratio$=\frac{\text{graph construction}}{\text{total training}}$.}
  \label{tab:Efficiency}
\end{table}

\section*{Conclusion}

We propose \textit{TSGDiff}, a novel framework for structured time series generation that rethinks the problem from a graph perspective. By unifying dynamic graph construction, Fourier spectrum-based edge modeling, and diffusion-based generative processes in latent graph space, \textit{TSGDiff} effectively captures intricate temporal dependencies and structural relationships. The model generates time series with realistic dynamics, high temporal accuracy, and structural coherence, addressing key challenges in traditional approaches. 
Additionally, we introduce the Topological Structure Fidelity (Topo-FID) score, a novel graph-aware metric that combines Graph Edit Similarity and Structural Entropy Similarity to comprehensively evaluate the quality of generated time series. This metric fills critical gaps in traditional evaluation methods by providing a more accurate and meaningful assessment of structural similarity. 
\textit{TSGDiff} offers a robust and flexible solution for synthetic time series generation, setting a foundation for future research into scalable, adaptive, and high-fidelity generative models.

}

\section{Acknowledgments}
This work was supported in part by the National Natural
Science Foundation of China under Grant Nos. 62221005,
62450043, 62222601, 62176033 and 62576056.

\bibliography{aaai2026}

@misc{c:23,
  title        = "Pluto: The 'other' red planet",
  author       = "{NASA}",
  howpublished = "\url{https://www.nasa.gov/nh/pluto-the-other-red-planet}",
  year         = 2015,
  note         = "Accessed: 2018-12-06"
}

@inproceedings{r:24x,
author = "Yuan, Xinyu and Qiao, Yan",
year = 2024,
title = "{Diffusion-TS: interpretable diffusion for general time series generation}",
booktitle = "ICLR"
}

@inproceedings{r:19x,
author = "Jinsung Yoon and Daniel Jarrett and Mihaela Van der Schaar",
year = 2019,
title = "{Time-series generative adversarial networks}",
booktitle = "NeurIPS",
volume = 32
}

@inproceedings{r:22x,
author = "Jeha Paul and Bohlke-Schneider Michael and Mercado Pedro and Kapoor Shubham and Singh Nirwan Rajbir and Flunkert Valentin and Gasthaus Jan and Januschowski Tim",
year = 2022,
title = "{Psa-gan: progressive self attention gans for synthetic time series}",
booktitle = "ICLR"
}

@article{r:08x,
author = "Laurens Van der Maaten and Geoffrey Hinton",
year = 2008,
title = "{Visualizing data using t-sne}",
journal = "Journal of Machine Learning Research",
volume = 9,
number = 11
}

@article{r:20x,
author = "Hao Ni and Lukasz Szpruch and Magnus Wiese and Shujian Liao and Baoren Xiao",
year = 2020,
title = "{Conditional sig-wasserstein gans for time series generation}",
journal = "Preprint arXiv:2006.05421"
}

@inproceedings{r:1a,
  title={TimeVAE: a variational auto-encoder for multivariate time series generation},
  author={Desai, Abhyuday and Freeman, Cynthia and Beaver, Ian and Wang, Zuhui},
  booktitle={ICLR},
  year={2022}
}

@inproceedings{r:2a,
  author = "Yusuke Tashiro and Jiaming Song and Yang Song and Stefano Ermon",
  year = 2021,
  title = "CSDI:conditional score-based diffusion models for probabilistic time series imputation",
  booktitle = "NeurIPS"
}

@inproceedings{r:4a,
  author    = {Haoyu Han and Mengdi Zhang and Min Hou and Fuzheng Zhang and Zhongyuan Wang and Enhong Chen and others},
  title     = {{STGCN}: A Spatial-temporal aware graph learning method for {POI} recommendation},
  booktitle = {ICDM},
  year      = {2020},
}

@inproceedings{r:5a,
  author    = {Kun Yi and Qi Zhang and Wei Fan and Hui He and Liang Hu and Pengyang Wang and Ning An and Longbing Cao and Zhendong Niu},
  title     = {Fourier graph neural network for multivariate time series forecasting},
  booktitle = {ICLR},
  year      = {2024}
}

@inproceedings{r:9a,
  title={Denoising diffusion probabilistic models},
  author={Ho, Jonathan and Jain, Ajay and Abbeel, Pieter},
  booktitle={NeurIPS},
  year={2020}
}

@article{r:10a,
    author = {Chen, Yongbao and Xu, Peng and Chu, Yiyi and Li, Weilin and Wu, Yuntao and Ni, Lizhou and Bao, Yi and Wang, Kun},
    title = {Short-term electrical load forecasting using the Support Vector Regression (SVR) model to calculate the demand response baseline for office buildings},
    journal = {Applied Energy},
    volume = {195},
    pages = {659--670},
    year = {2017}
}

@article{r:11a,
    author = {Bao, Wei and Yue, Jun and Rao, Yulei},
    title = {A deep learning framework for financial time series using stacked autoencoders and long-short term memory},
    journal = {PLoS ONE},
    volume = {12},
    number = {7},
    year = {2017}
}

@article{r:12a,
author = {Zammel, Zina and Khabou, Nesrine and Souifi, Lotfi and Rodriguez, Ismael Bouassida},
title = {Time series prediction models in healthcare: Systematic literature review},
journal = {ReDCAD Laboratory, ENIS, University of Sfax, Tunisia},
year = {2023},
}

@article{r:13a,
author = {Esteban, Cristobal and Hyland, Stephanie L. and Ratsch, Gunnar},
title = {Real-valued (medical) time series generation with recurrent conditional GANs},
journal = {Preprint arXiv:1706.02633},
year = {2017}
}

@article{r:14a,
author = {Fortuin, Vincent and Baranchuk, Dmitry and Ratsch, Gunnar and Mandt, Stephan},
title = {GP-VAE: deep probabilistic time series imputation},
journal = {AISTATS},
year = {2020},
volume = {108},
pages = {1651--1661}
}

@article{r:15a,
author = {Shen, Lifeng and Kwok, James T.},
title = {Non-autoregressive conditional diffusion models for time series prediction},
journal = {ICML},
year = {2023},
volume = {202},
pages = {30512--30525}
}

@book{r:86x,
author = "Ronald Newbold Bracewell and Ronald N Bracewell",
year = 1986,
title = "{The fourier transform and its applications}",
publisher = "McGraw - Hill New York",
volume = 31999
}

@inproceedings{r:23x,
  author  = "Kun Yi and Qi Zhang and Wei Fan and Hui He and Liang Hu and Pengyang Wang and Ning An and Longbing Cao and Zhendong Niu",
  year    = 2023,
  title   = "{FourierGNN: rethinking multivariate time series forecasting from a pure graph perspective}",
  booktitle = "NeurIPS"
}

@inproceedings{r:17x,
  author    = "Thomas N. Kipf and Max Welling",
  year      = 2017,
  title     = "{Semi-supervised classification with graph convolutional networks}",
  booktitle = "ICLR",
}

@book{r:70x,
  author    = "Y. Katznelson",
  year      = 1970,
  title     = "{An introduction to harmonic analysis}",
  publisher = "Cambridge University Press",
}

@article{s:17x,
  author  = "Santiago Segarra and Gonzalo Mateos and Antonio G. Marques and Alejandro Ribeiro",
  year    = 2017,
  title   = "{Blind identification of graph filters}",
  journal = "IEEE Trans. Signal Process.",
  volume  = 65,
  number  = 5,
  pages   = "1146--1159"
}

@article{e:21x,
  author  = "Isufi, Elvin and Gama, Fernando and Ribeiro, Alejandro",
  year    = 2021,
  title   = "{EdgeNets: edge varying graph neural networks}",
  journal = "IEEE",
  volume = 44,
  pages    = "7457--7473"
}

@inproceedings{kieu2019outlier,
  title={Outlier detection for time series with recurrent autoencoder ensembles},
  author={Kieu, Tung and Yang, Bin and Guo, Chenjuan and Jensen, Christian S},
  booktitle={IJCAI},
  year={2019}
}

@inproceedings{shen2021time,
  title={Time series anomaly detection with multiresolution ensemble decoding},
  author={Shen, Lifeng and Yu, Zhongzhong and Ma, Qianli and Kwok, James T},
  booktitle={AAAI},
  year={2021}
}

@article{yu2016temporal,
  title={Temporal regularized matrix factorization for high-dimensional time series prediction},
  author={Yu, Hsiang-Fu and Rao, Nikhil and Dhillon, Inderjit S},
  journal={NeurIPS},
  year={2016}
}

@article{song2020denoising,
  title={Denoising diffusion implicit models},
  author={Song, Jiaming and Meng, Chenlin and Ermon, Stefano},
  journal={ICLR},
  year={2021}
}

@inproceedings{cheng2020time2graph,
  title={Time2graph: revisiting time series modeling with dynamic shapelets},
  author={Cheng, Ziqiang and Yang, Yang and Wang, Wei and Hu, Wenjie and Zhuang, Yueting and Song, Guojie},
  booktitle={AAAI},
  year={2020}
}

@article{paszke2019pytorch,
author = "Paszke, Adam and Gross, Sam and Massa, Francisco and Lerer, Adam and Bradbury, James and Chanan, Gregory and Killeen, Trevor and Lin, Zeming and Gimelshein, Natalia and Antiga, Luca and Desmaison, Alban and Köpf, Andreas and Yang, Edward and DeVito, Zach and Raison, Martin and Tejani, Alykhan and Chilamkurthy, Sasank and Steiner, Benoit and Fang, Lu and Bai, Junjie and Chintala, Soumith",
year = 2019,
title = "{PyTorch: an imperative style, high-performance deep learning library}",
journal = "NeurIPS",
volume = 32
}

@inproceedings{naiman2024utilizing,
  title={Utilizing image transforms and diffusion models for generative modeling of short and long time series},
  author={Naiman, Ilan and Berman, Nimrod and Pemper, Itai and Arbiv, Idan and Fadlon, Gal and Azencot, Omri},
  booktitle={NeurIPS},
  year={2024},
}

@article{crabbé2024time,
  title={Time series diffusion in the frequency domain},
  author={Crabbé, Jonathan and Huynh, Nicolas and Stanczuk, Jan and van der Schaar, Mihaela},
  journal={ICML},
  year={2024}
}

@article{zhou2023deep,
  title={Deep latent state space models for time-series generation},
  author={Zhou, Linqi and Poli, Michael and Xu, Winnie and Massaroli, Stefano and Ermon, Stefano},
  journal={ICML},
  year={2023}
}

@article{park2024leveraging,
  title={Leveraging priors via diffusion bridge for time series generation},
  author={Park, Jinseong and Lee, Seungyun and Jeong, Woojin and Choi, Yujin and Lee, Jaewook},
  journal={Preprint arXiv:2408.06672v1},
  year={2024}
}

@inproceedings{alaa2021generative,
  title={Generative time-series modeling with fourier flows},
  author={Alaa, Ahmed M. and Chan, Alex J. and van der Schaar, Mihaela},
  booktitle={ICLR},
  year={2021}
}

\clearpage
\onecolumn 
\appendix

\section{Supplementary Algorithms}
The core workflow of TSGDiff comprises two main stages: training and sampling, with their respective implementations detailed in Algorithms \ref{alg:tsgdiff} and \ref{alg:tsgdiff_sampling}.

i) In the \textit{Training} stage (Algorithm \ref{alg:tsgdiff}), the process begins by preprocessing time series data and constructing graph structures. A joint optimization is then performed using an encoder-decoder framework integrated with a diffusion model. The objective function combines the weighted sum of reconstruction loss, KL divergence loss, denoising loss, and Fourier loss.

ii) In the \textit{Sampling} stage (Algorithm \ref{alg:tsgdiff_sampling}), the process starts with standard normal noise, which is iteratively denoised through a reverse diffusion process guided by the trained model. The decoder ultimately generates synthetic time series that align with the original distribution characteristics.

\begin{algorithm}
\caption{Training of \textit{TSGDiff} Model}
\label{alg:tsgdiff}
\textbf{Input:} Time series $X$; Window size $W$; Diffusion steps $K$; \\
\qquad \quad Model params ($H, E$); Epochs $N$ \\
\textbf{Output:} Trained TSGDiff model $\theta$

\begin{algorithmic}[1]
\REPEAT
  \STATE Preprocess $X$: normalize to $[-1,1]$; slice into $S = \{s_i\}$ with size $W$
  \STATE For each $s_i$:
  \STATE \qquad Extract frequency features via FFT; detect top-3 periods
  \STATE \qquad Generate graph adjacency matrix $A_i$ from periods
  \STATE Initialize model with Encoder $E$ (GraphConv + residual), Decoder $D$ (Linear + Mish), \\
  \qquad Diffusion model $F$ (time-step embedded denoising)
  \FOR{epoch = 1 to $N$}
    \FOR{batch in training data}
      \STATE Sample $t \sim \text{Uniform}(0, K)$; encode $s_i$ to $z$ (get $\mu, \log\sigma$)
      \STATE Reparameterize: $z_{\text{sample}} = \mu + \epsilon \exp(\log\sigma/2), \epsilon \sim \mathcal{N}(0,I)$
      \STATE Add noise: $z_t = \sqrt{\alpha_t} z_{\text{sample}} + \sqrt{1-\alpha_t} \epsilon$
      \STATE Predict $\hat{z} = F(z_t, t, A_i)$; reconstruct $\hat{s}_i = D(\hat{z})$
      \STATE Compute losses:
      \STATE \qquad $\mathcal{L}_{\text{recon}} = \text{MSE}(\hat{s}_i, s_i)$; $\mathcal{L}_{\text{KL}} = -0.5\mathbb{E}[1 + 2\log\sigma - \mu^2 - \exp(2\log\sigma)]$
      \STATE \qquad $\mathcal{L}_{\text{diff}} = \text{MSE}(\hat{z}, z_{\text{sample}})$; $\mathcal{L}_{\text{fourier}} = \text{MSE}(\text{FFT}(\hat{s}_i), \text{FFT}(s_i))$
      \STATE \qquad $\mathcal{L}_{\text{total}} = \mathcal{L}_{\text{recon}} + \lambda_1\mathcal{L}_{\text{KL}} + \lambda_2\mathcal{L}_{\text{diff}} + \lambda_3\mathcal{L}_{\text{fourier}}$
      \STATE Update $\theta$ via gradient descent on $\mathcal{L}_{\text{total}}$
    \ENDFOR
  \ENDFOR
\UNTIL{Model converges}

\STATE \textbf{Return} $\theta$
\end{algorithmic}
\end{algorithm}

\begin{algorithm}
\caption{Sampling of \textit{TSGDiff} Model}
\label{alg:tsgdiff_sampling}
\textbf{Input:} Trained $\theta$; Samples $N$; Dim $E, V, D$; Diffusion steps $K$ \\
\textbf{Output:} Synthetic time series $\hat{X}$

\begin{algorithmic}[1]
\STATE Set $\theta$ to evaluation mode; sample $z_0 \sim \mathcal{N}(0, I_{E \times E})$ for $N$ samples
\FOR{$t = K-1$ down to $0$}
  \STATE $t_{\text{batch}} = \text{full}(N, t)$; predict $\hat{z} = \theta_{\text{diffusion}}(z_t, t_{\text{batch}})$
  \STATE Get precomputed $\alpha_t, \alpha_{\text{cum}}, \alpha_{\text{prev}}$ ($1.0$ if $t=0$)
  \STATE $\beta_t = \theta_{\beta}[t]$; $\sigma_t^2 = \beta_t \cdot (1 - \alpha_{\text{prev}}) / (1 - \alpha_{\text{cum}})$
  \STATE Update: $z_{t-1} = \sqrt{1/\alpha_t} \cdot \left(z_t - \frac{1 - \alpha_t}{\sqrt{1 - \alpha_{\text{cum}}}} \cdot (z_t - \hat{z})\right)$
  \STATE If $t > 0$: add $\epsilon \sim \mathcal{N}(0, I)$: $z_{t-1} += \sqrt{\sigma_t^2} \cdot \epsilon$
\ENDFOR
\STATE Decode: $\hat{X} = \theta_{\text{decoder}}(z_{-1})$; apply tanh to $[-1, 1]$
\STATE \textbf{Return} $\hat{X}$
\end{algorithmic}
\end{algorithm}

\newpage
\section{Supplementary Experimental Results}

\subsection{Effects of Time Series Length}
To further assess the stability of our model in generating long multivariate time series, we conducted additional experiments on the \texttt{ETTh} dataset, generating multivariate time series of varying lengths.

\subsubsection{Results for Synthetic Long-term Time Series Generation}
Table \ref{tab:add} presents the generation results of \textit{TSGDiff}, Diffusion-TS, TimeGAN, Cot-GAN, and TimeVAE for time series of lengths 64, 96, and 128. The results demonstrate that \textit{TSGDiff} consistently outperforms all baselines across nearly all metrics, producing synthetic samples of superior quality. Notably, \textit{TSGDiff} excels in the Context-FID score, maintaining excellent performance across all sequence lengths. Unlike other baselines, the performance of \textit{TSGDiff} remains remarkably stable as sequence length increases. This highlights its superior long-term robustness, making it well-suited for practical applications that involve synthesizing complex, long time series.

\begin{table*}[h!]
  \centering
  \begin{tabular}{ccccccc} 
    \midrule[1pt]
    Metric & Length & TSGDiff & Diffusion-TS & TimeGAN & Cot-GAN & TimeVAE \\
    \midrule
    \multirow{3}{*}{\makecell[c]{Topo-FID $\nearrow$ \\ (\footnotesize{higher the better)}}} 
    & 64 & \textbf{0.875$\pm$\textsubscript{0.0\%}} & 0.823$\pm$\textsubscript{0.0\%} & 0.785$\pm$\textsubscript{0.0\%} & \underline{0.859$\pm$\textsubscript{0.0\%}} & 0.821$\pm$\textsubscript{0.0\%} \\
    & 96 & \textbf{0.952$\pm$\textsubscript{0.0\%}} & \underline{0.937$\pm$\textsubscript{0.0\%}} & 0.794$\pm$\textsubscript{0.0\%} & 0.906$\pm$\textsubscript{0.0\%} & 0.913$\pm$\textsubscript{0.0\%} \\
    & 128 & \textbf{0.848$\pm$\textsubscript{0.0\%}} & \underline{0.836$\pm$\textsubscript{0.0\%}} & 0.805$\pm$\textsubscript{0.0\%} & 0.818$\pm$\textsubscript{0.0\%} & 0.822$\pm$\textsubscript{0.0\%} \\

    \midrule
    \multirow{3}{*}{\makecell[c]{Context-FID $\searrow$ \\ \footnotesize (lower the better)}} 
    & 64 & \textbf{0.306$\pm$\textsubscript{0.5\%}} & \underline{0.487$\pm$\textsubscript{0.9\%}} & 6.181$\pm$\textsubscript{0.6\%} & 3.675$\pm$\textsubscript{0.7\%} & 3.038$\pm$\textsubscript{1.2\%} \\
    & 96 & \textbf{0.367$\pm$\textsubscript{0.3\%}} & \underline{0.412$\pm$\textsubscript{0.9\%}} & 11.631$\pm$\textsubscript{1.1\%} & 3.823$\pm$\textsubscript{0.8\%} & 3.476$\pm$\textsubscript{2.1\%} \\
    & 128 & \textbf{0.428$\pm$\textsubscript{0.6\%}} & \underline{1.031$\pm$\textsubscript{0.5\%}} & 7.877$\pm$\textsubscript{0.8\%} & 7.716$\pm$\textsubscript{0.9\%} & 3.787$\pm$\textsubscript{1.5\%} \\

    \midrule
    \multirow{3}{*}{\makecell[c]{Correlational $\searrow$ \\ \footnotesize (lower the better)}}  
    & 64 & \textbf{0.024$\pm$\textsubscript{0.1\%}} & 0.030$\pm$\textsubscript{0.2\%} & 0.319$\pm$\textsubscript{0.3\%} & 0.137$\pm$\textsubscript{0.3\%} & \underline{0.027$\pm$\textsubscript{0.4\%}} \\
    & 96 & \textbf{0.022$\pm$\textsubscript{0.1\%}} & 0.027$\pm$\textsubscript{0.3\%} & 0.567$\pm$\textsubscript{0.5\%} & 0.117$\pm$\textsubscript{0.2\%} & \underline{0.022$\pm$\textsubscript{0.2\%}} \\
    & 128 & \textbf{0.021$\pm$\textsubscript{0.2\%}} & 0.047$\pm$\textsubscript{0.4\%} & 0.489$\pm$\textsubscript{0.2\%} & 0.144$\pm$\textsubscript{0.3\%} & \underline{0.023$\pm$\textsubscript{0.3\%}} \\

    \midrule
    \multirow{3}{*}{\makecell[c]{Discriminative $\searrow$ \\ \footnotesize (lower the better)}} 
    & 64 & \textbf{0.110$\pm$\textsubscript{0.3\%}} & 0.181$\pm$\textsubscript{0.5\%} & 0.448$\pm$\textsubscript{1.6\%} & \underline{0.140$\pm$\textsubscript{1.5\%}} & 0.374$\pm$\textsubscript{1.3\%} \\
    & 96 & \textbf{0.086$\pm$\textsubscript{0.2\%}} & 0.226$\pm$\textsubscript{0.7\%} & 0.497$\pm$\textsubscript{0.7\%} & \underline{0.172$\pm$\textsubscript{1.0\%}} & 0.402$\pm$\textsubscript{0.5\%} \\
    & 128 & \textbf{0.104$\pm$\textsubscript{0.4\%}} & 0.349$\pm$\textsubscript{0.9\%} & 0.492$\pm$\textsubscript{0.8\%} & \underline{0.118$\pm$\textsubscript{1.3\%}} & 0.407$\pm$\textsubscript{0.6\%} \\

    \midrule
    \multirow{3}{*}{\makecell[c]{Predictive $\searrow$ \\ \footnotesize (lower the better)}}  
    & 64 & \textbf{0.020$\pm$\textsubscript{0.0\%}} & \underline{0.025$\pm$\textsubscript{0.0\%}} & 0.034$\pm$\textsubscript{0.1\%} & 0.088$\pm$\textsubscript{0.2\%} & 0.145$\pm$\textsubscript{0.0\%} \\
    & 96 & \textbf{0.021$\pm$\textsubscript{0.0\%}} & 0.027$\pm$\textsubscript{0.2\%} & \underline{0.026$\pm$\textsubscript{0.1\%}} & 0.090$\pm$\textsubscript{0.1\%} & 0.146$\pm$\textsubscript{0.2\%} \\
    & 128 & \textbf{0.020$\pm$\textsubscript{0.1\%}} & \underline{0.025$\pm$\textsubscript{0.1\%}} & 0.029$\pm$\textsubscript{0.2\%} & 0.086$\pm$\textsubscript{0.1\%} & 0.190$\pm$\textsubscript{0.2\%} \\

    \bottomrule
  \end{tabular}
  \caption{Detailed Results of Long-term Time-series Generation on \texttt{ETTh} dataset. (bold indicates best performance, while underline is the second best).}
  \label{tab:add}
\end{table*}

\subsubsection{Visualization Results}
To further illustrate the high quality of the time series generated by \textit{TSGDiff}, we employed t-SNE, PCA, and kernel density estimation analyses to visualize how well the distribution of generated data aligns with that of real data.

\begin{figure*}[htbp]
    \centering
    \setlength{\tabcolsep}{0pt}
    \hspace*{-\tabcolsep}
    
    \begin{minipage}[b]{0.37\textwidth}
        \centering
        \includegraphics[width=\textwidth, trim=0 0 0 0, clip]{figures/ETTh1t.png}\\
        \includegraphics[width=\textwidth, trim=0 0 0 0, clip]{figures/ETTh0t.png}
        \caption*{(a) \texttt{ETTh}-48}
    \end{minipage}
    \quad
    \begin{minipage}[b]{0.37\textwidth}
        \centering
        \includegraphics[width=\textwidth, trim=0 0 0 0, clip]{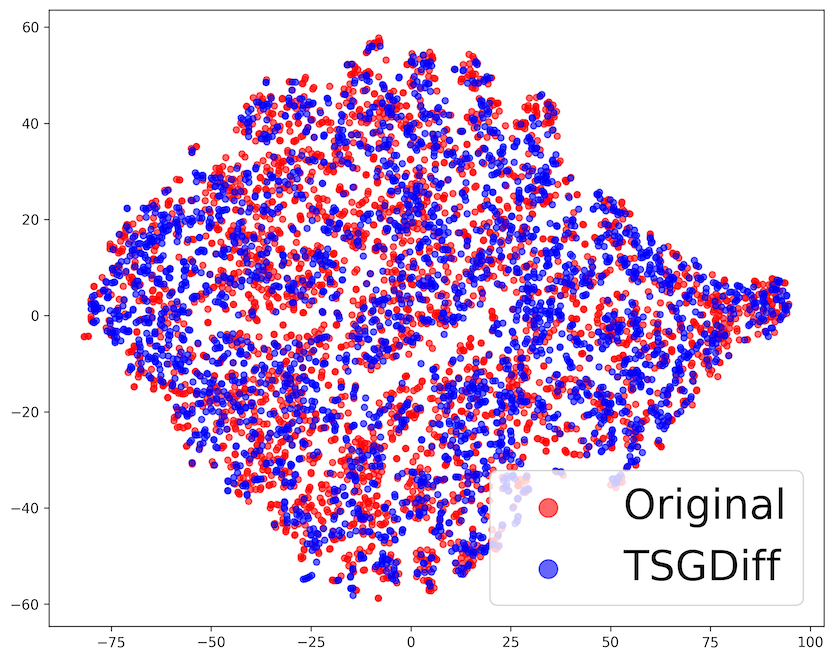}
        \includegraphics[width=\textwidth, trim=0 0 0 0, clip]{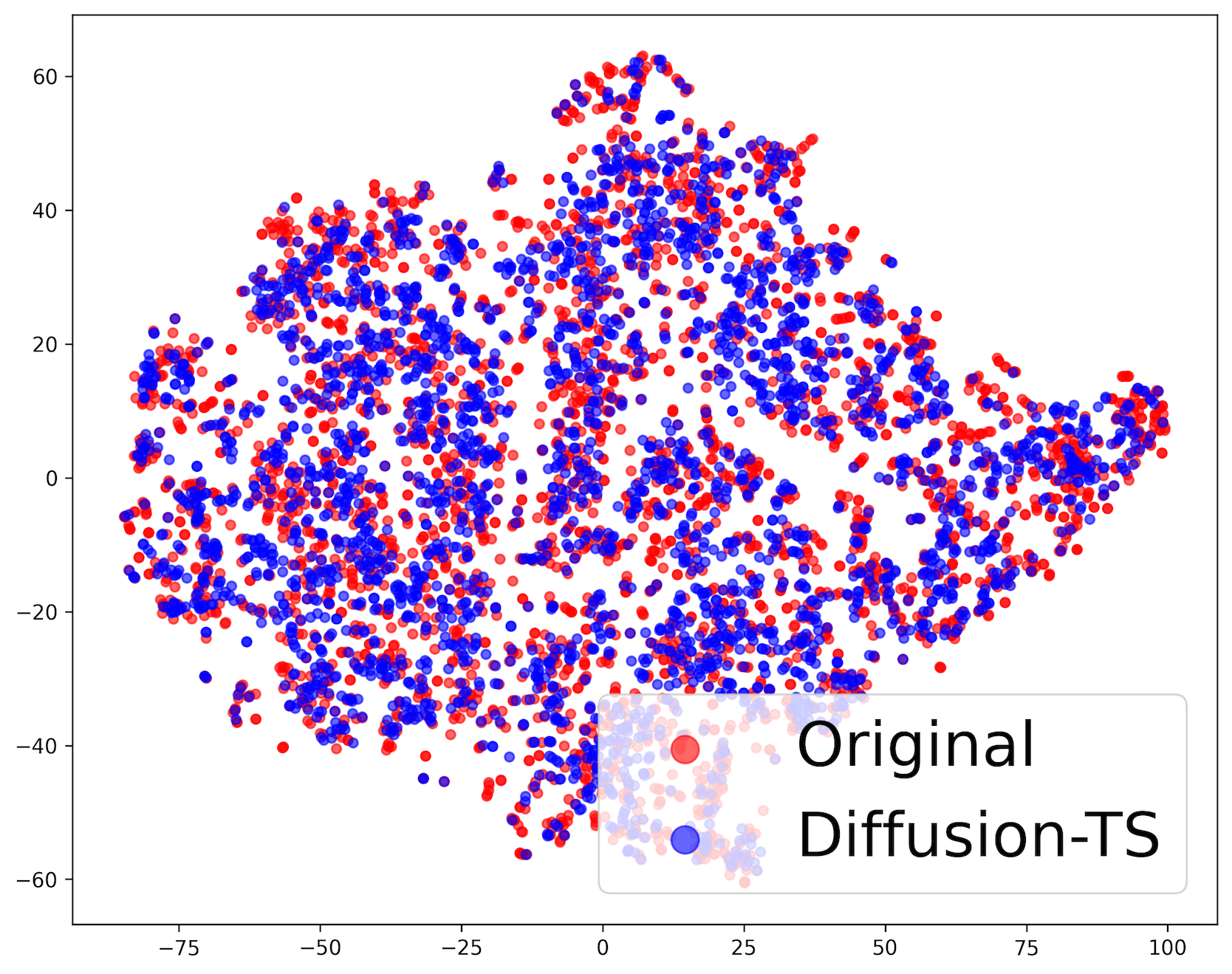}
        \caption*{(b) \texttt{ETTh}-64}
    \end{minipage}%
    \quad
    \begin{minipage}[b]{0.37\textwidth}
        \centering
        \includegraphics[width=\textwidth, trim=0 0 0 0, clip]{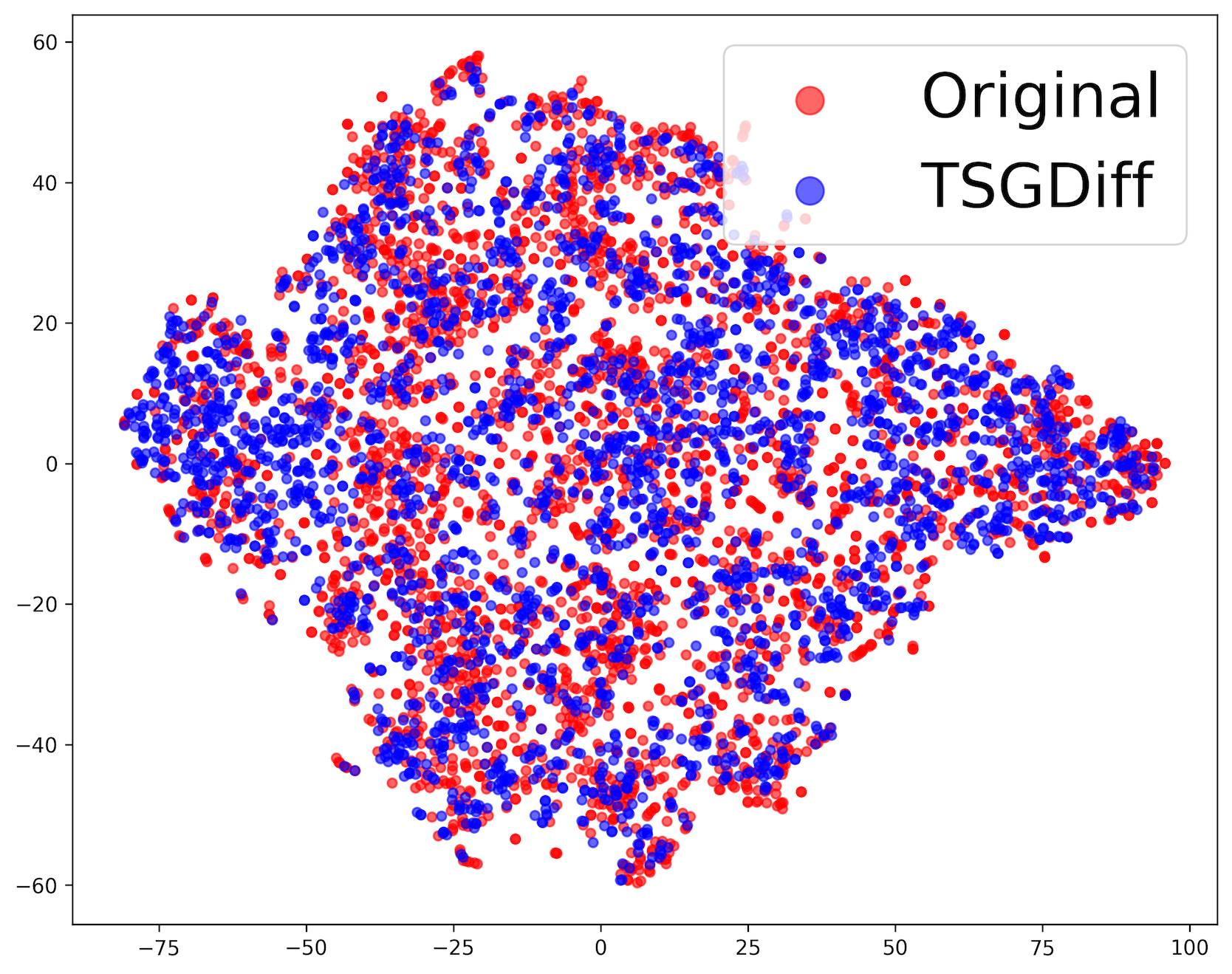}
        \includegraphics[width=\textwidth, trim=0 0 0 0, clip]{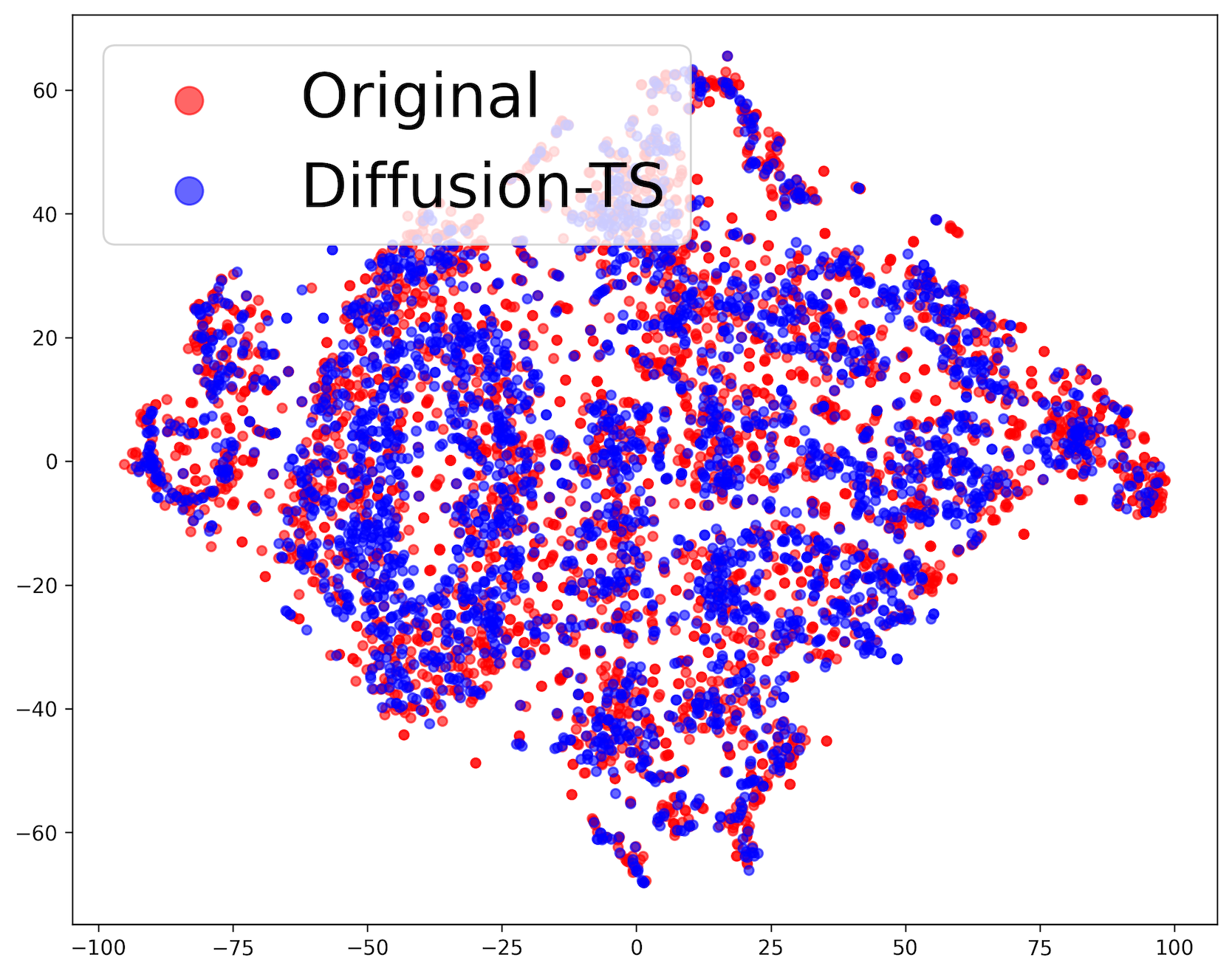}
        \caption*{(c) \texttt{ETTh}-96}
    \end{minipage}%
    \quad
    \begin{minipage}[b]{0.37\textwidth}
        \centering
        \includegraphics[width=\textwidth, trim=0 0 0 0, clip]{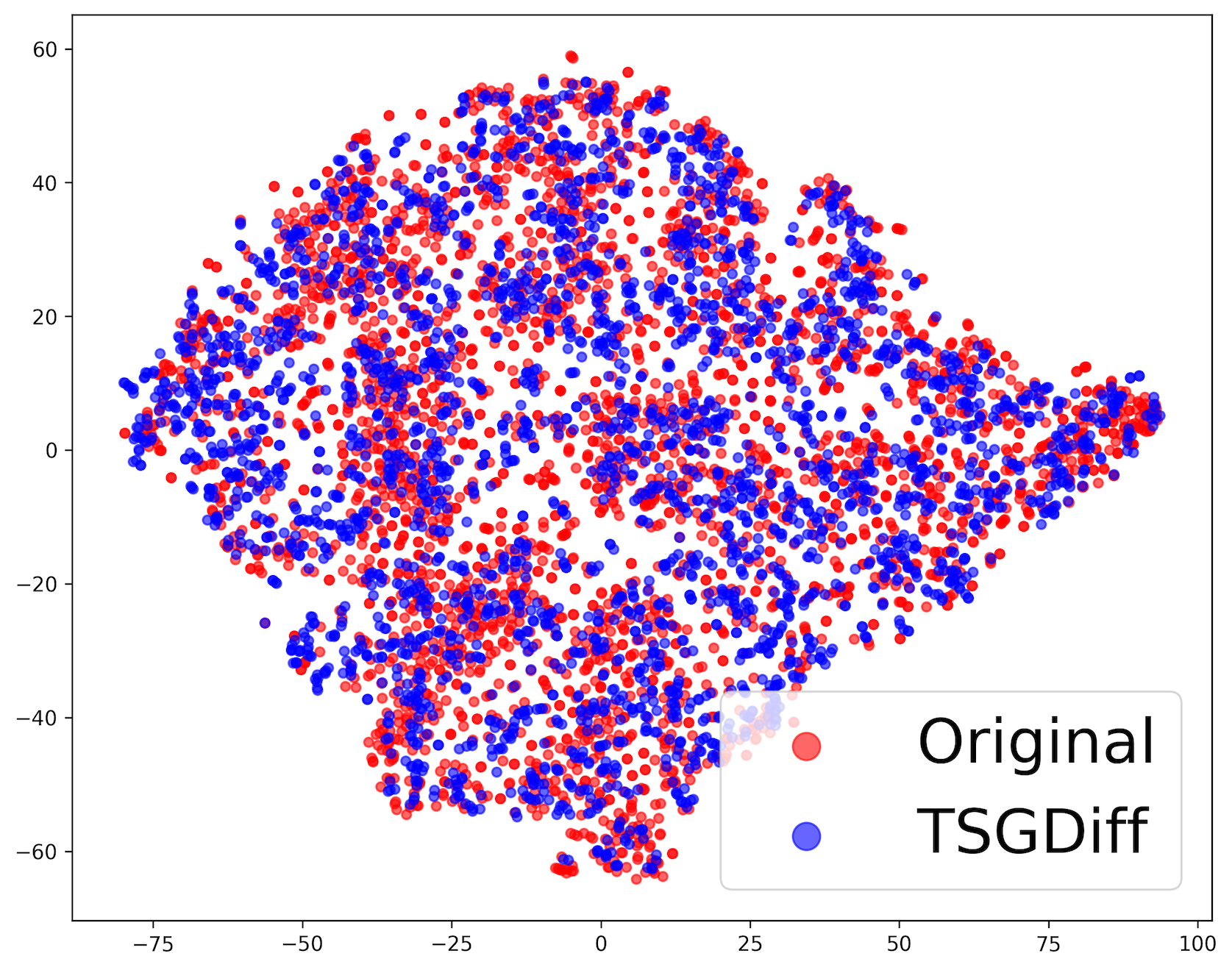}
        \includegraphics[width=\textwidth, trim=0 0 0 0, clip]{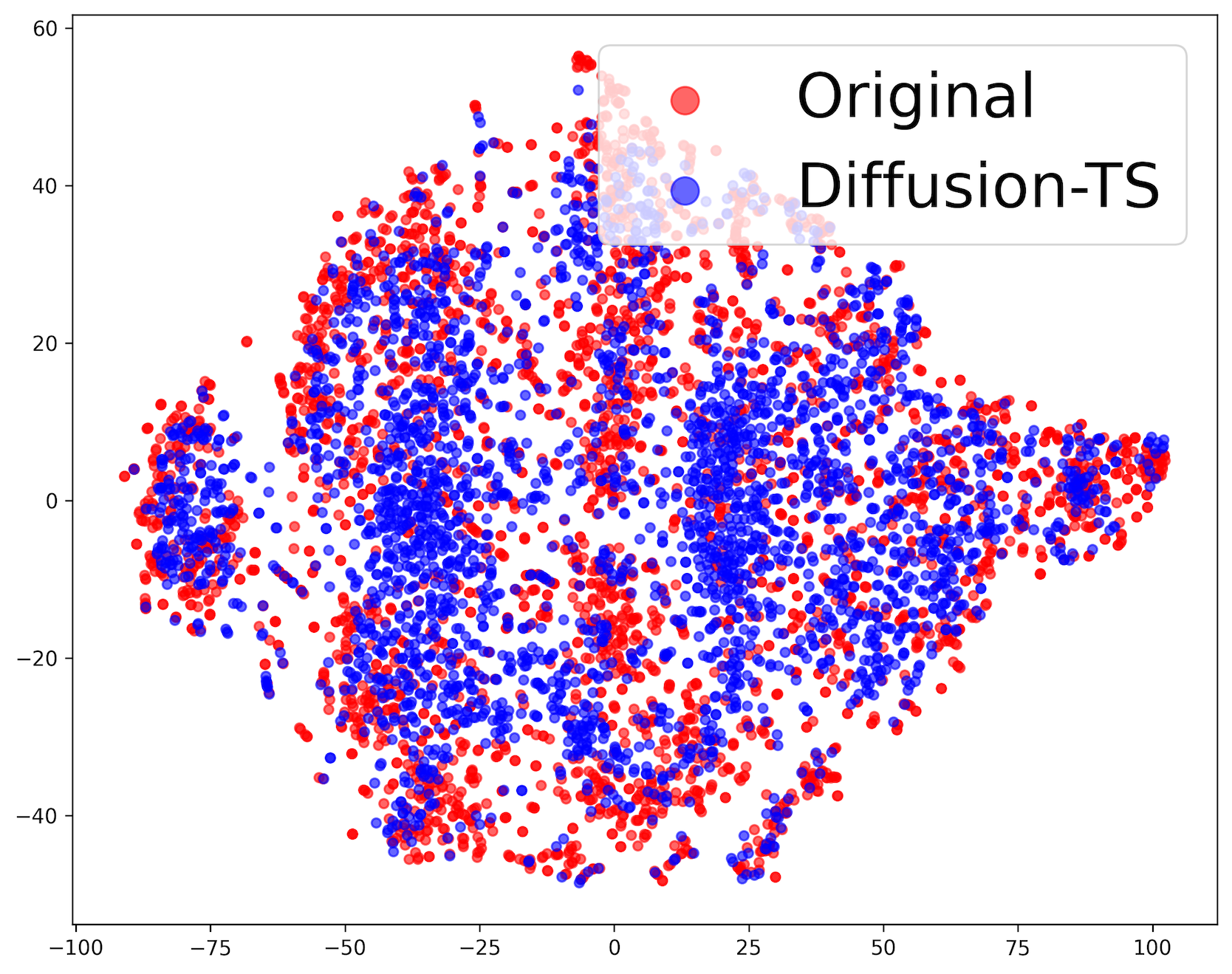}
        \caption*{(d) \texttt{ETTh}-128}
    \end{minipage}
    \caption{t-SNE plots of the time series of length 24, 64, 128 and 256 synthesized by \textit{TSGDiff} and Diffusion-TS. Red dots represent real data instances, and blue dots represent generated data samples in all plots.}
    \label{fig:tsnelong}
\end{figure*}

\begin{figure*}[htbp]
    \centering
    \setlength{\tabcolsep}{0pt}
    \hspace*{-\tabcolsep}
    
    \begin{minipage}[b]{0.37\textwidth}
        \centering
        \includegraphics[width=\textwidth, trim=0 0 0 0, clip]{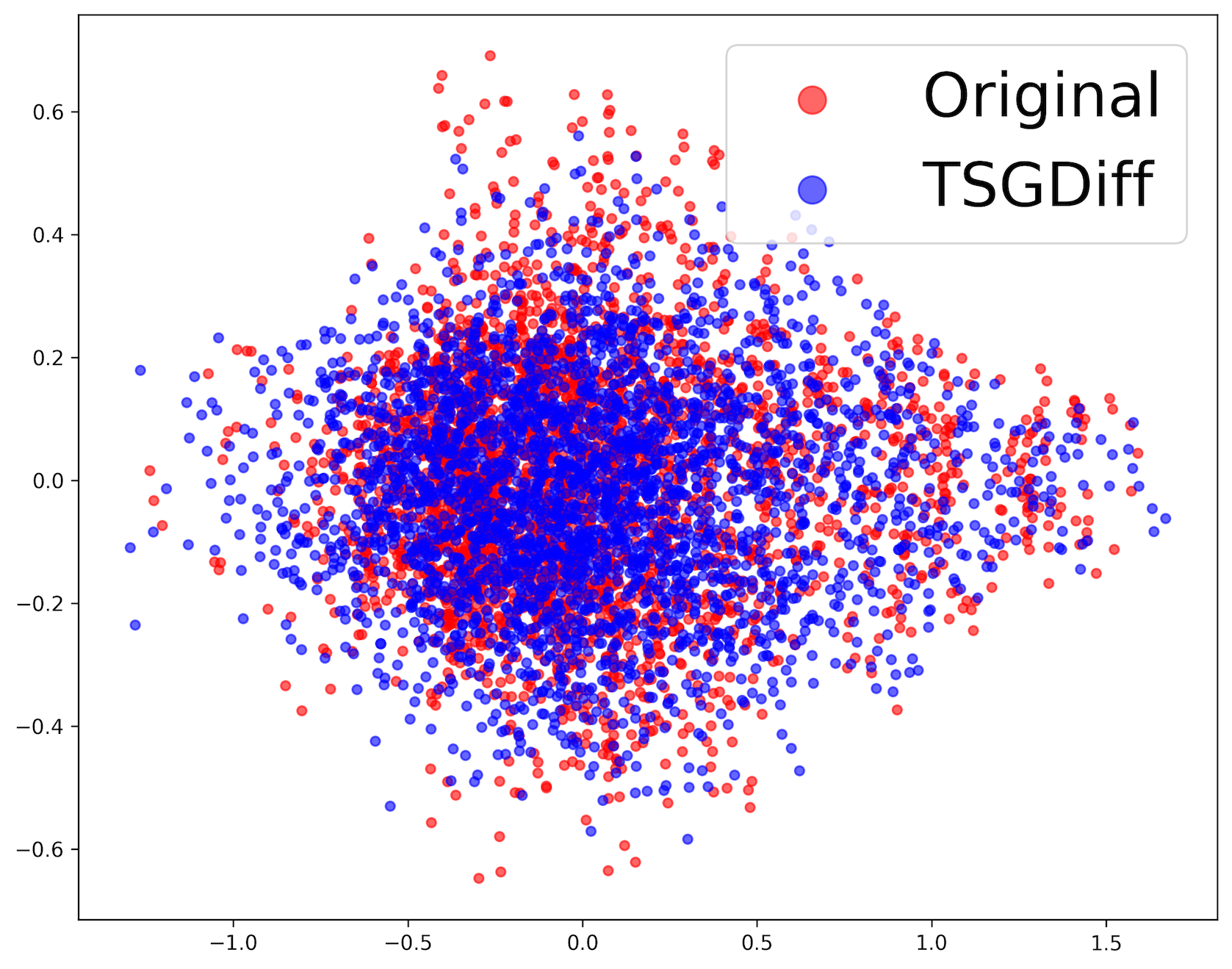}\\
        \includegraphics[width=\textwidth, trim=0 0 0 0, clip]{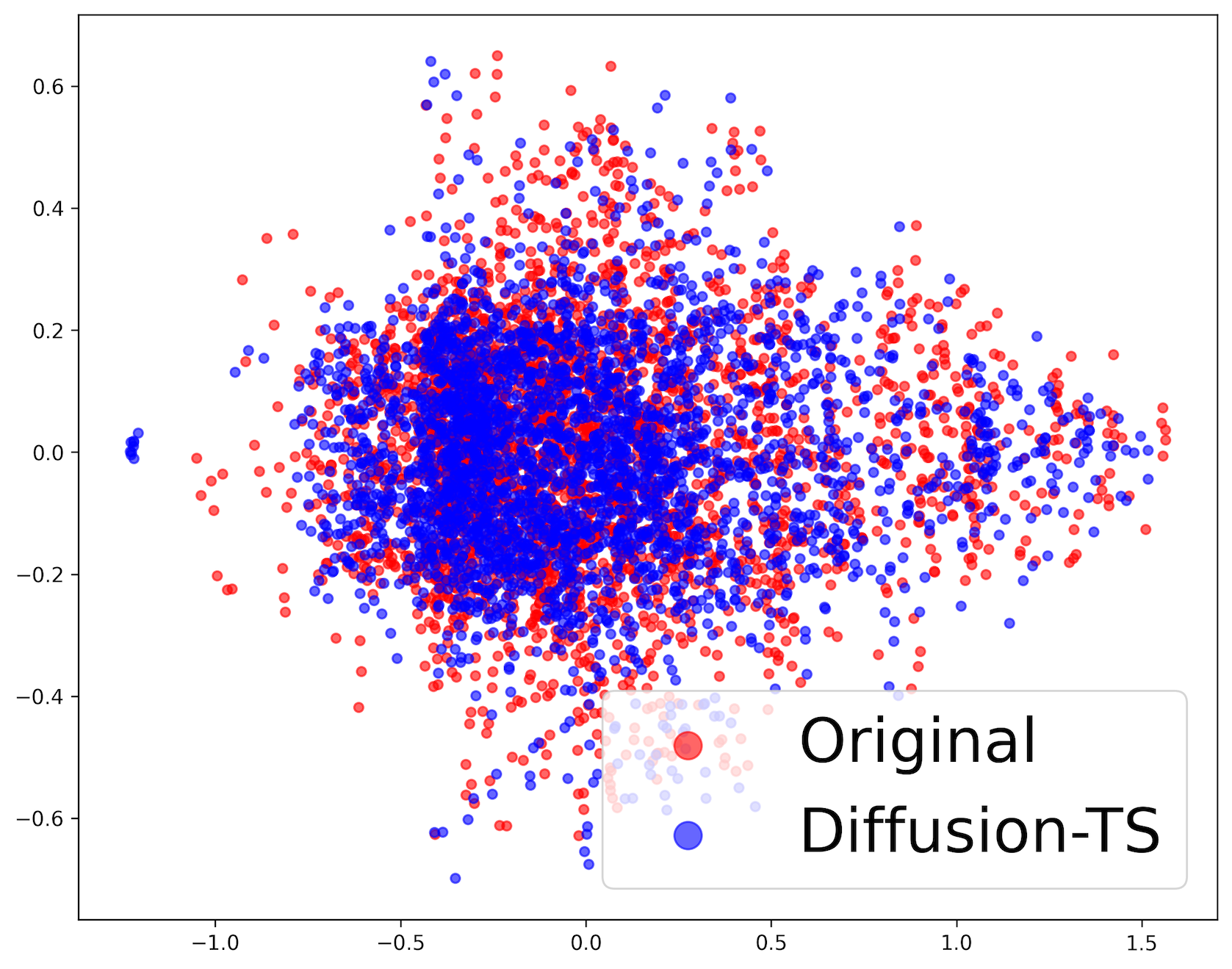}
        \caption*{(a) \texttt{ETTh}-48}
    \end{minipage}
    \quad
    \begin{minipage}[b]{0.37\textwidth}
        \centering
        \includegraphics[width=\textwidth, trim=0 0 0 0, clip]{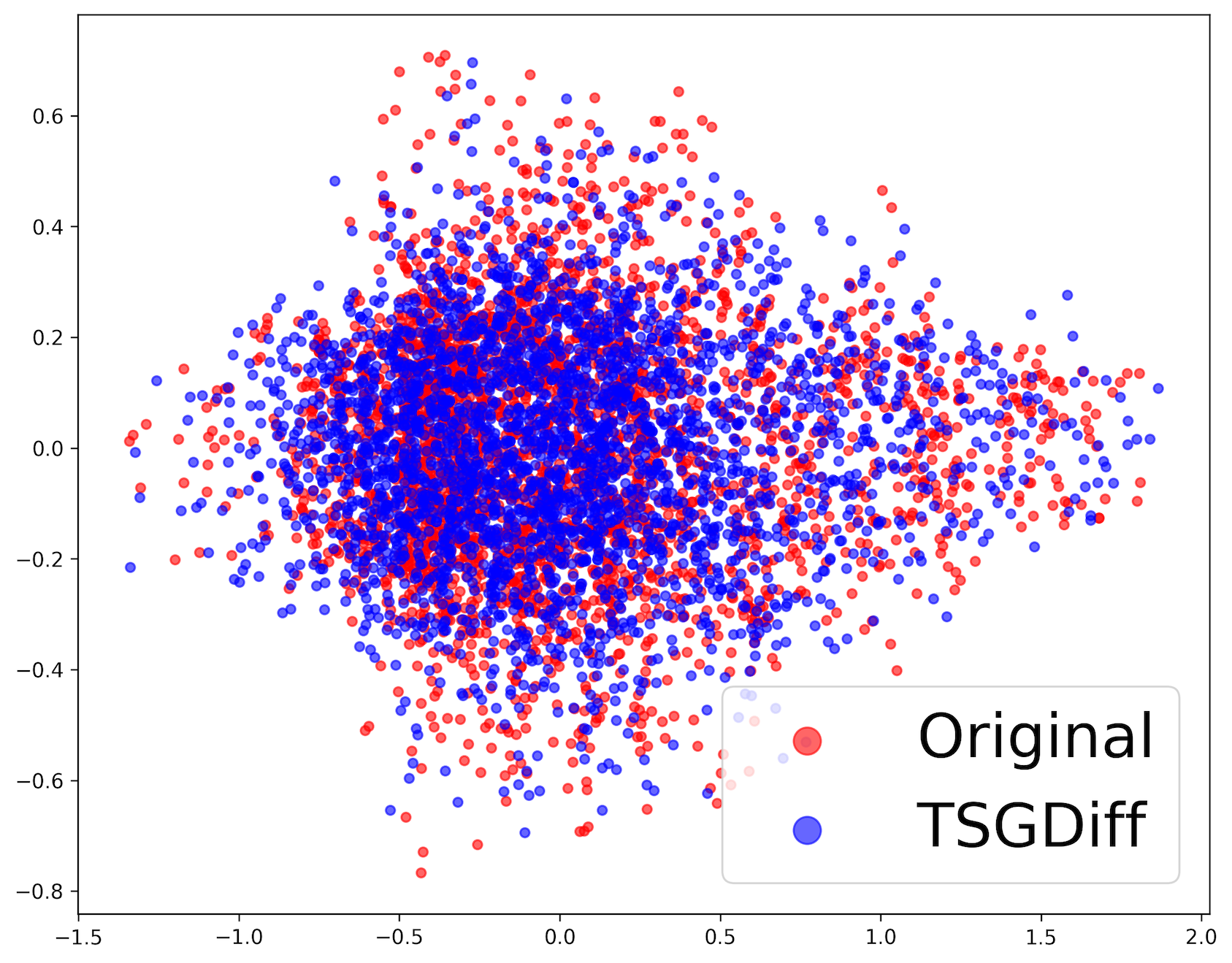}
        \includegraphics[width=\textwidth, trim=0 0 0 0, clip]{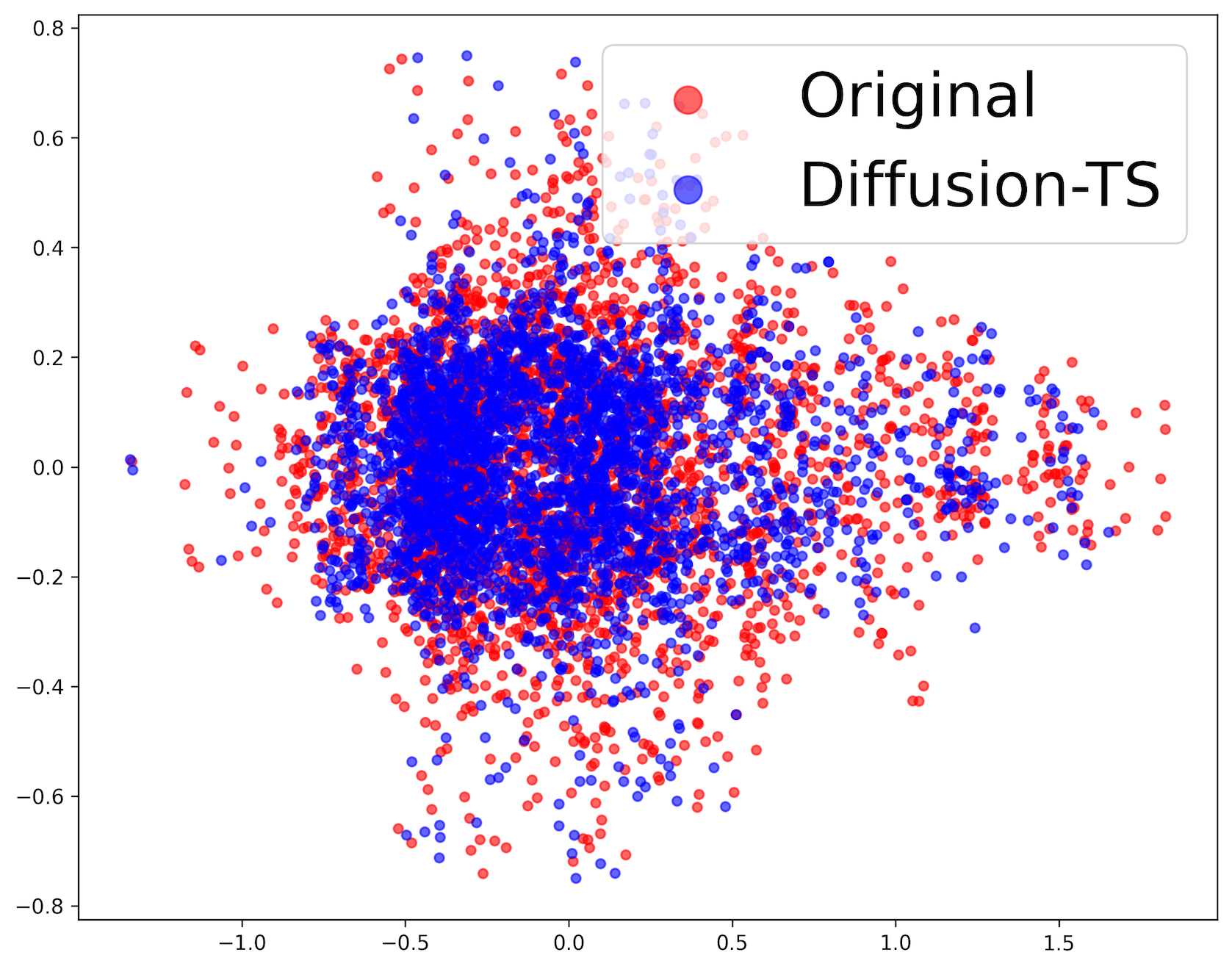}
        \caption*{(b) \texttt{ETTh}-64}
    \end{minipage}%
    \quad
    \begin{minipage}[b]{0.37\textwidth}
        \centering
        \includegraphics[width=\textwidth, trim=0 0 0 0, clip]{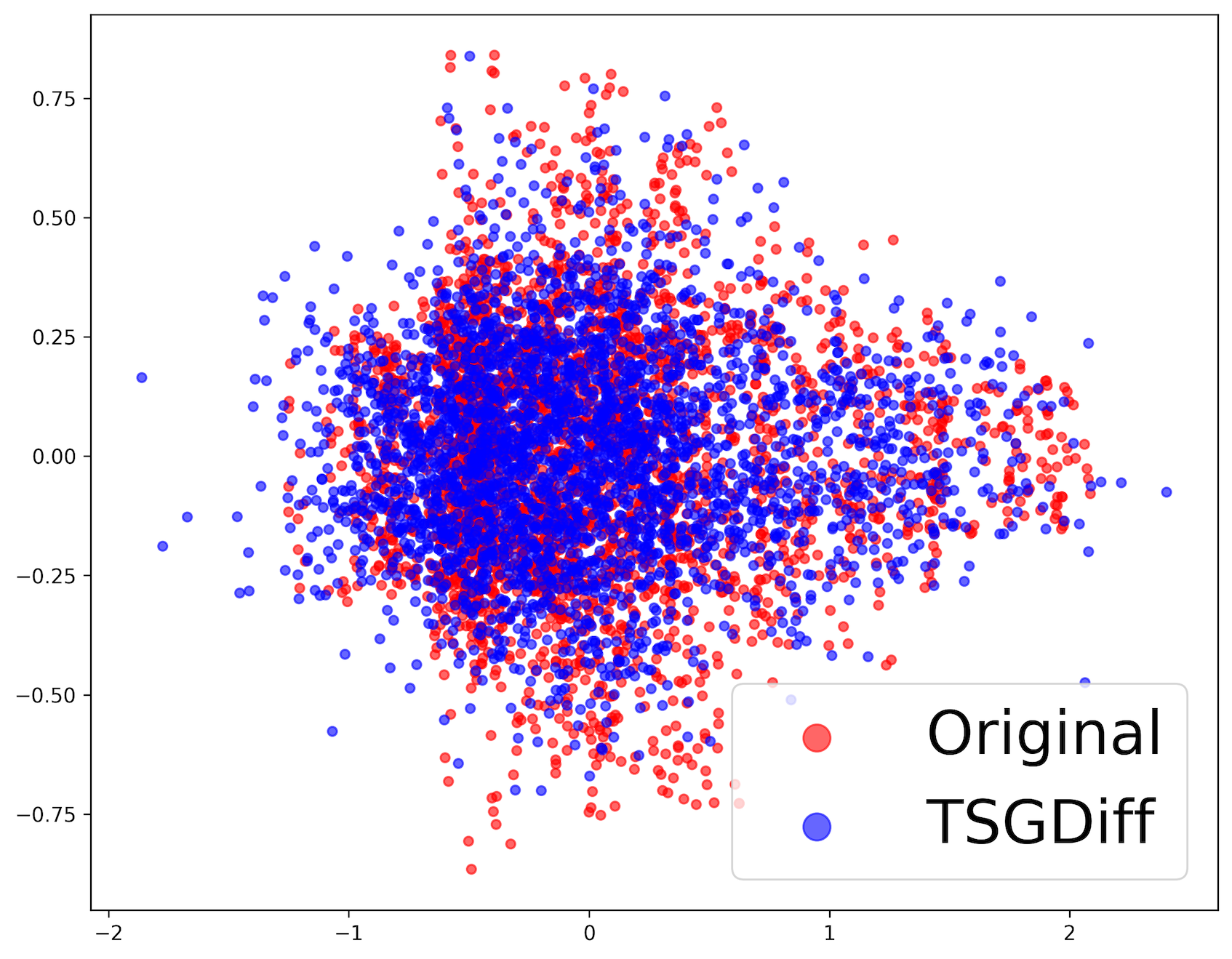}
        \includegraphics[width=\textwidth, trim=0 0 0 0, clip]{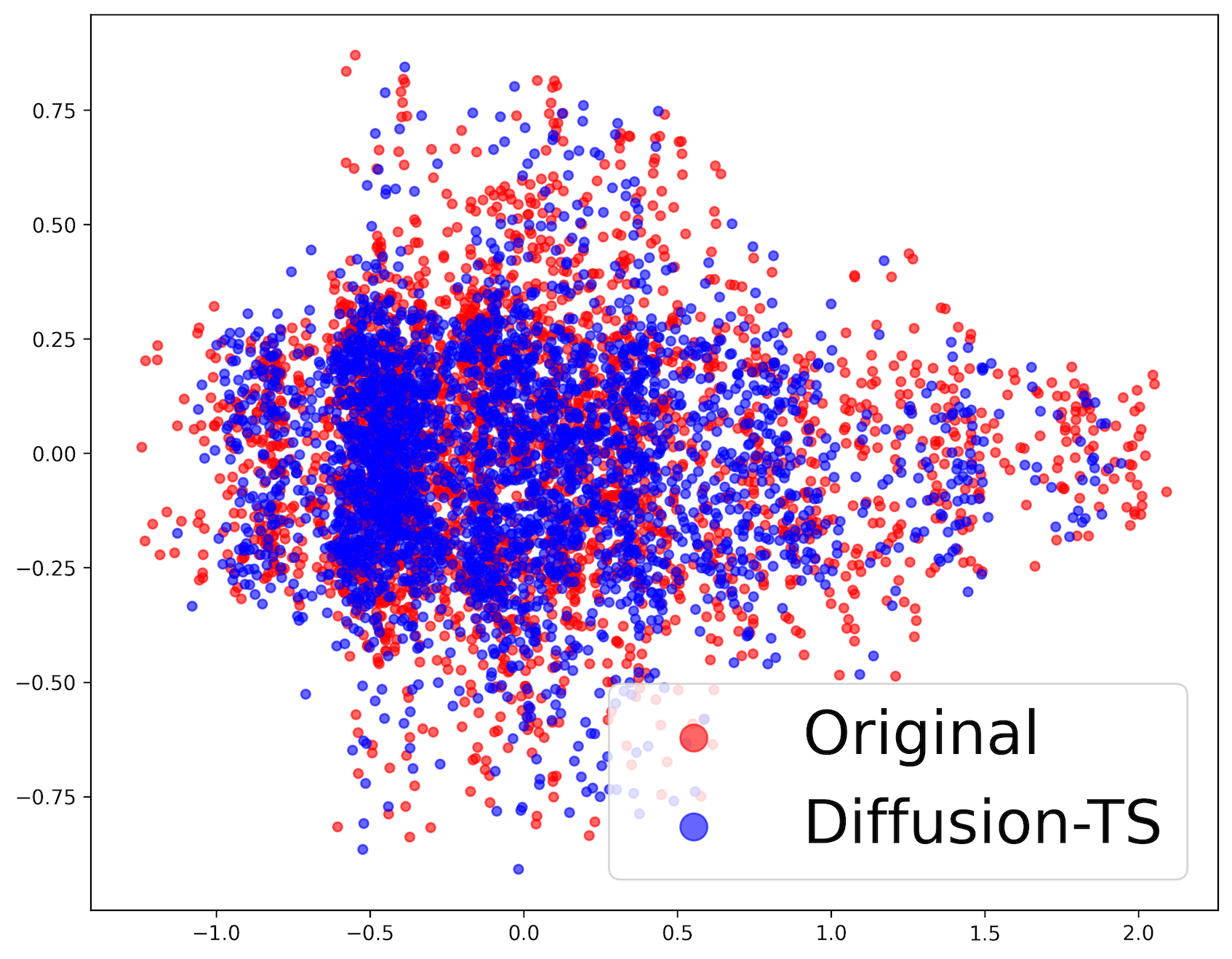}
        \caption*{(c) \texttt{ETTh}-96}
    \end{minipage}%
    \quad
    \begin{minipage}[b]{0.37\textwidth}
        \centering
        \includegraphics[width=\textwidth, trim=0 0 0 0, clip]{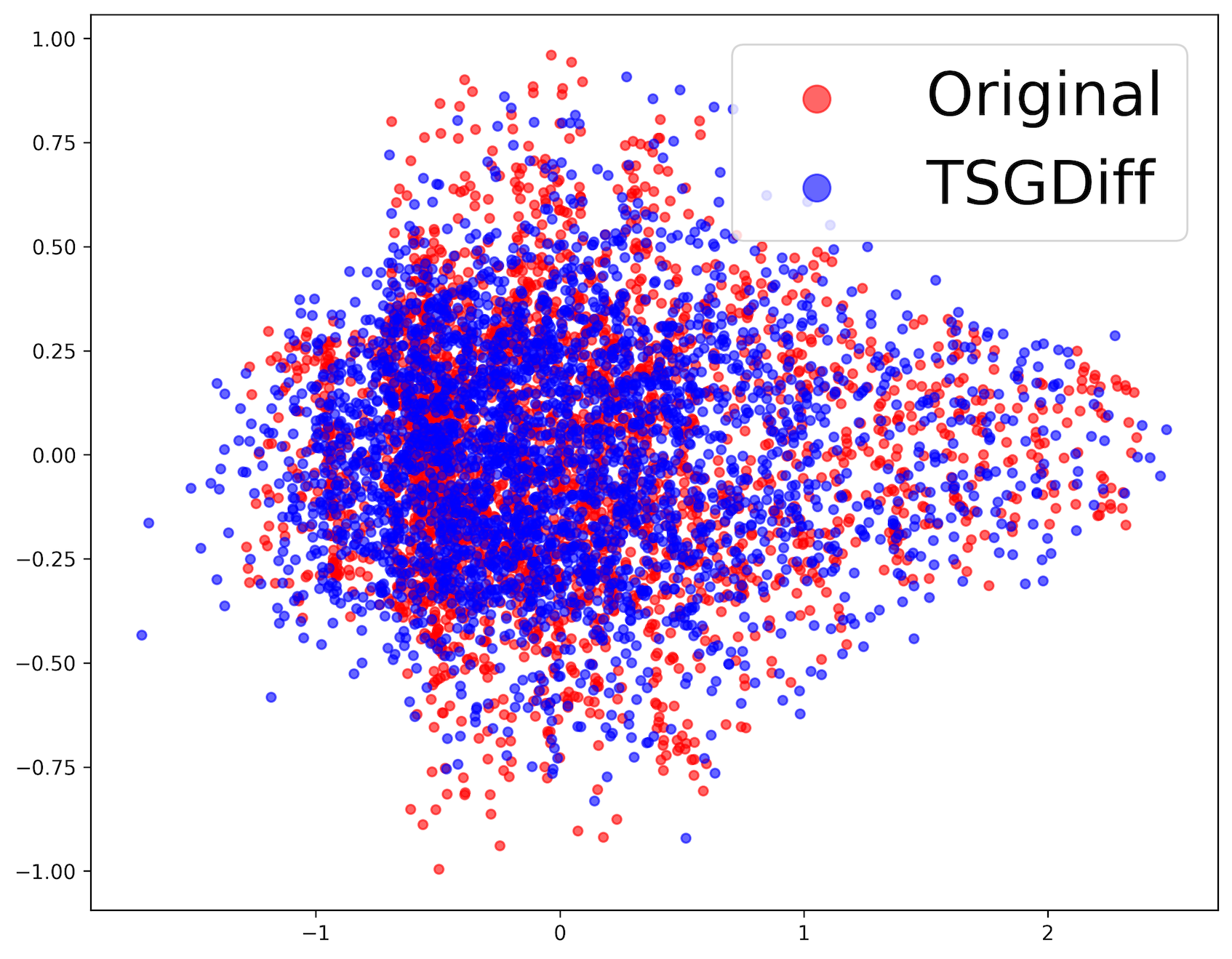}
        \includegraphics[width=\textwidth, trim=0 0 0 0, clip]{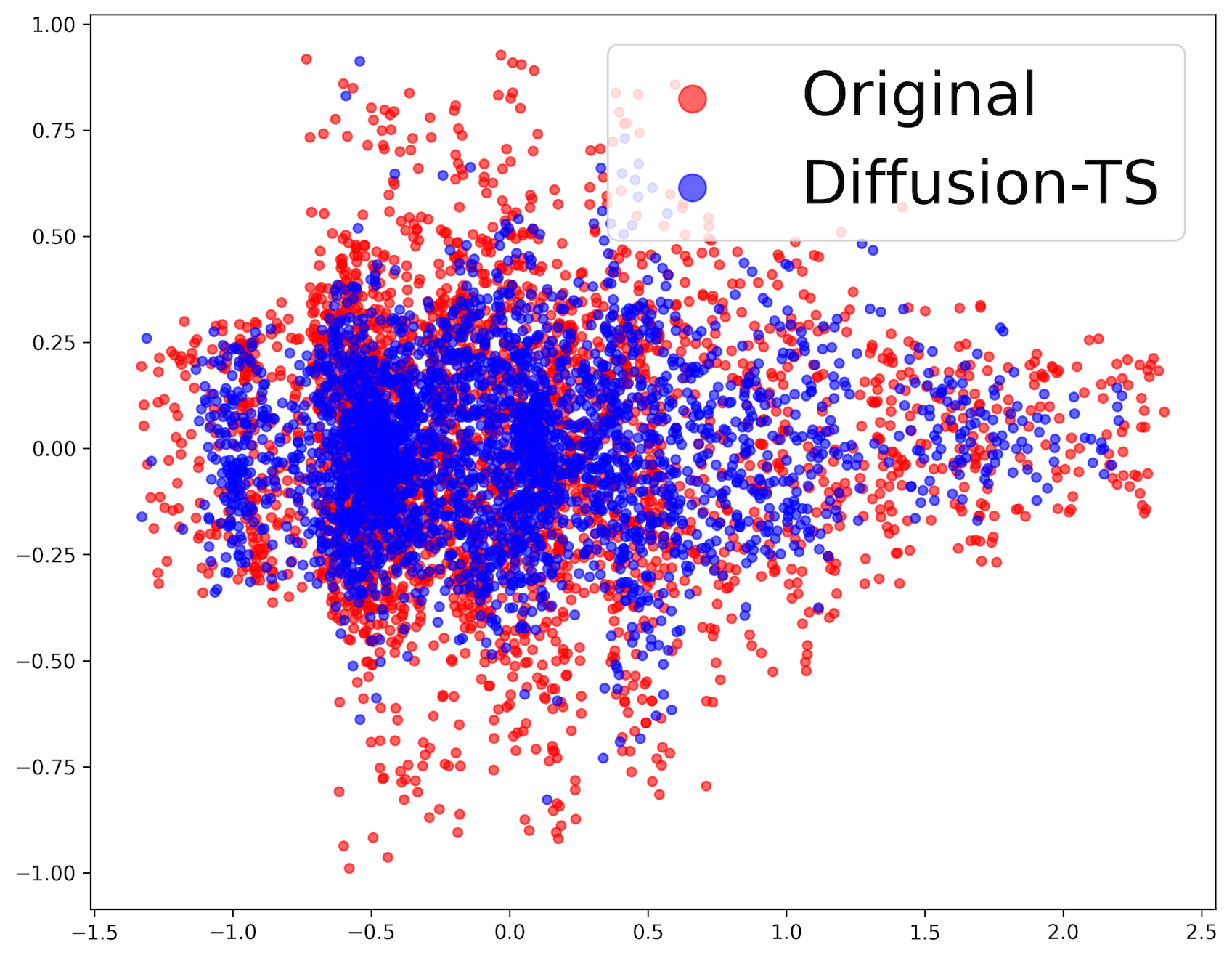}
        \caption*{(d) \texttt{ETTh}-128}
    \end{minipage}
    \caption{PCA plots of the time series of length 24, 64, 128 and 256 synthesized by \textit{TSGDiff} and Diffusion-TS. Red dots represent real data instances, and blue dots represent generated data samples in all plots.}
    \label{fig:pcalong}
\end{figure*}

\textbf{t-SNE and PCA Visualizations:}
Figures \ref{fig:tsnelong} and \ref{fig:pcalong} show the results for generating time series of lengths 48, 64, 96 and 128. Both \textit{TSGDiff} and Diffusion-TS exhibit substantial overlap with the original data in these visualizations. However, as the sequence length increases, \textit{TSGDiff} demonstrates exceptional stability, consistently generating data that closely aligns with the original distribution—an ability that Diffusion-TS fails to maintain.

\begin{figure*}[htbp]
    \centering
    \setlength{\tabcolsep}{0pt}
    \hspace*{-\tabcolsep}
    
    \begin{minipage}[b]{0.35\textwidth}
        \centering
        \includegraphics[width=\textwidth, trim=0 0 0 0, clip]{figures/ETTh1kk.png}\\
        \includegraphics[width=\textwidth, trim=0 0 0 0, clip]{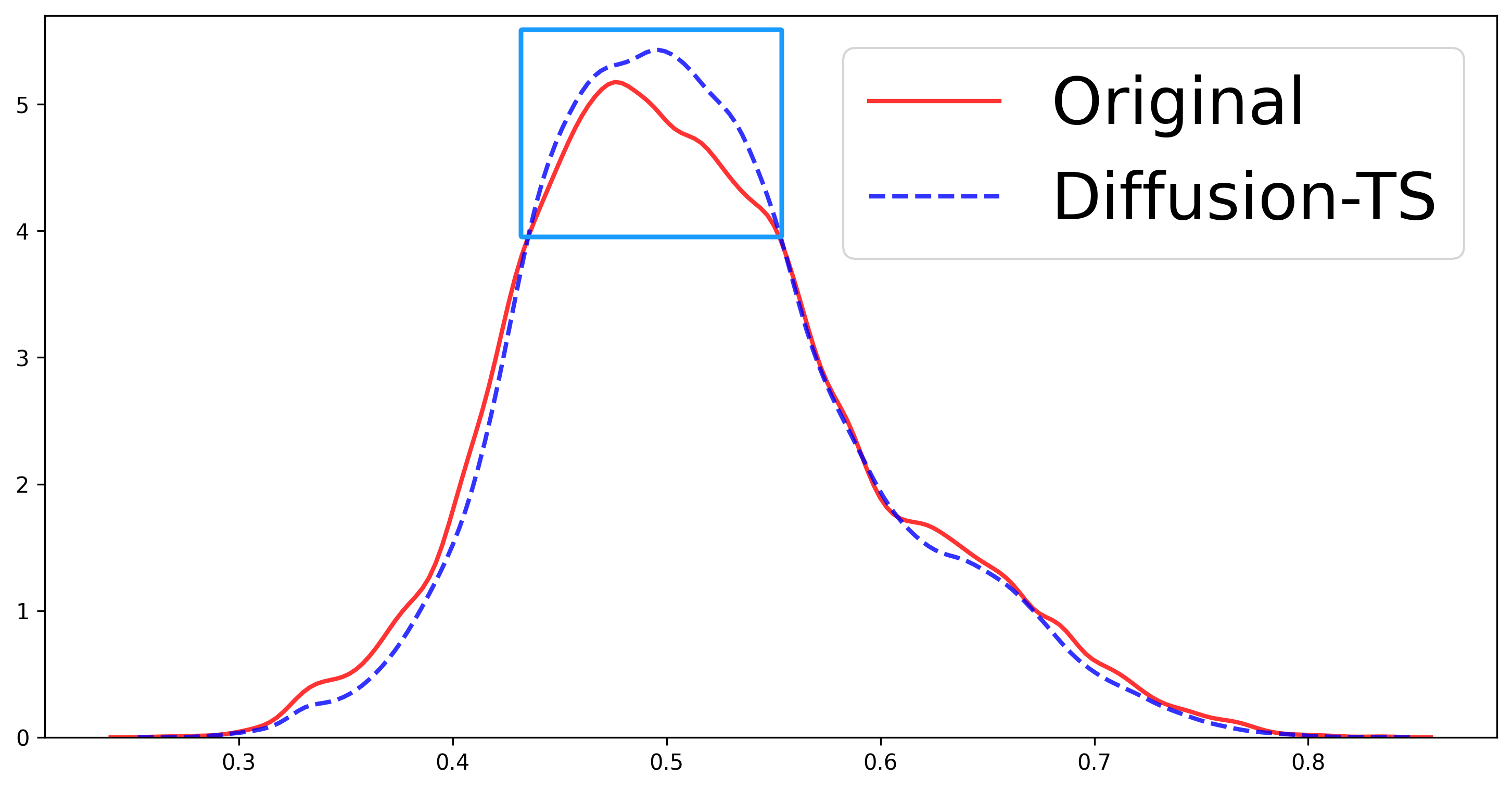}
        \caption*{(a) \texttt{ETTh}-48}
    \end{minipage}
    \quad
    \begin{minipage}[b]{0.35\textwidth}
        \centering
        \includegraphics[width=\textwidth, trim=0 0 0 0, clip]{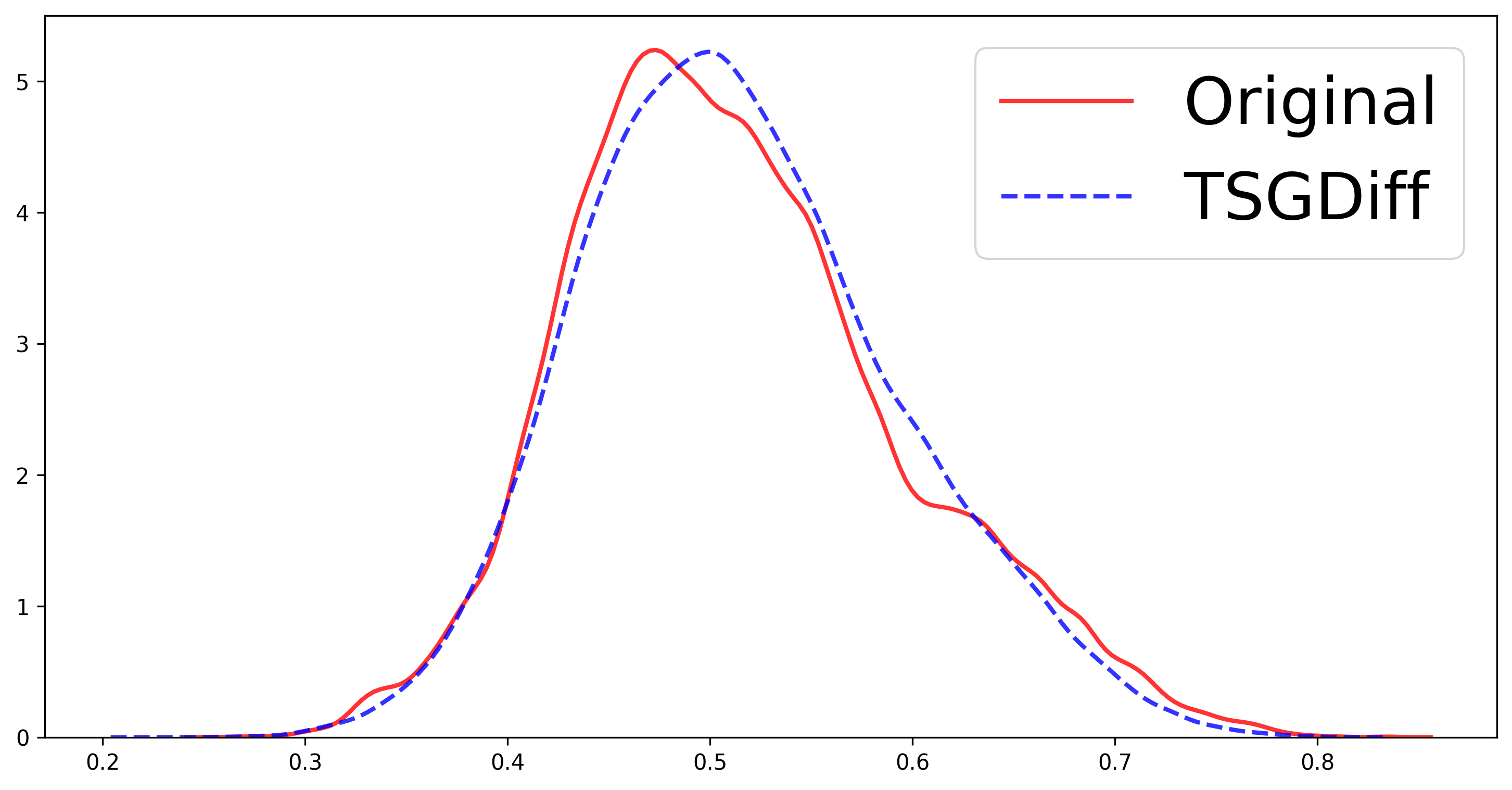}
        \includegraphics[width=\textwidth, trim=0 0 0 0, clip]{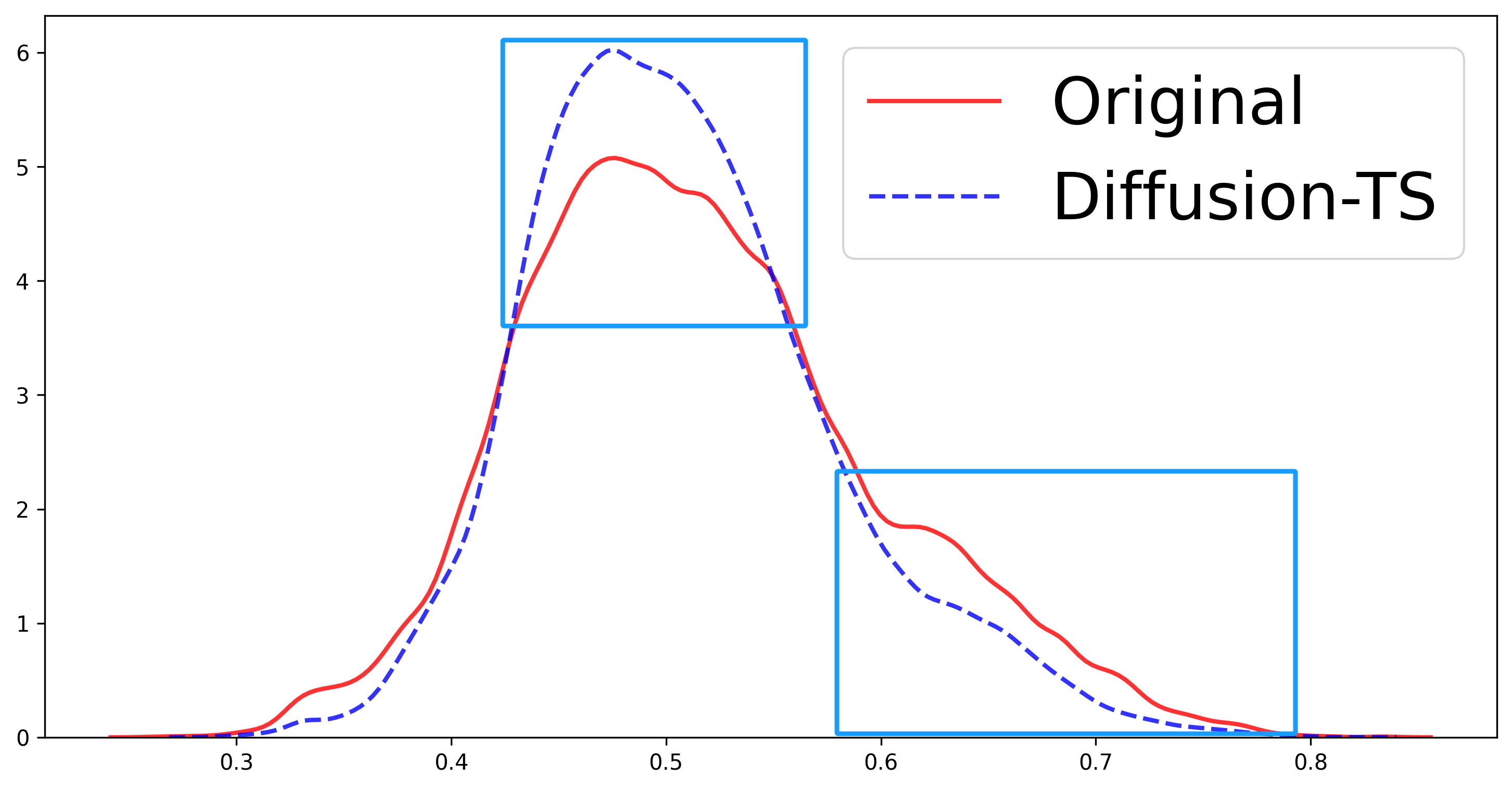}
        \caption*{(b) \texttt{ETTh}-64}
    \end{minipage}%
    \quad
    \begin{minipage}[b]{0.35\textwidth}
        \centering
        \includegraphics[width=\textwidth, trim=0 0 0 0, clip]{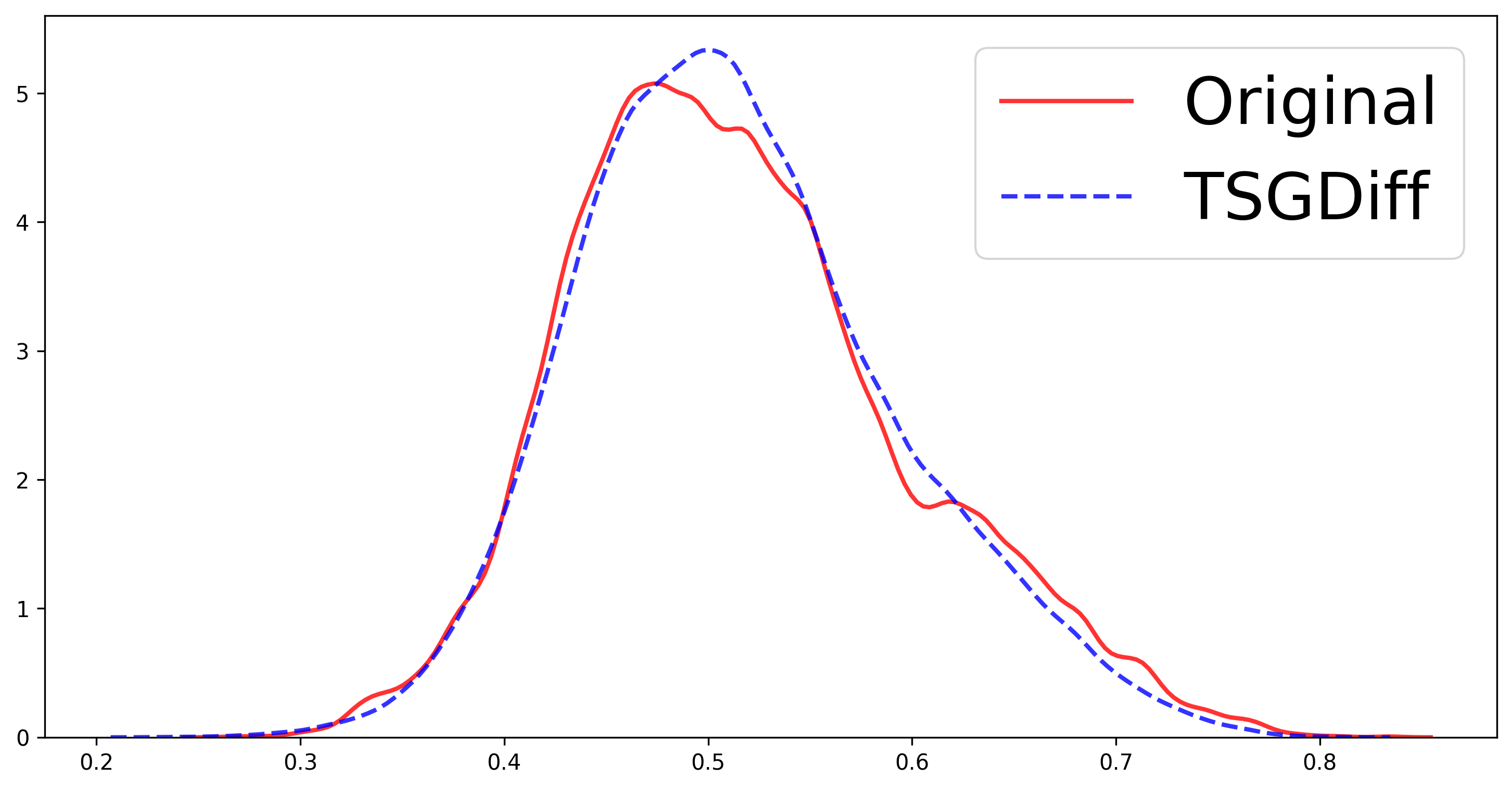}
        \includegraphics[width=\textwidth, trim=0 0 0 0, clip]{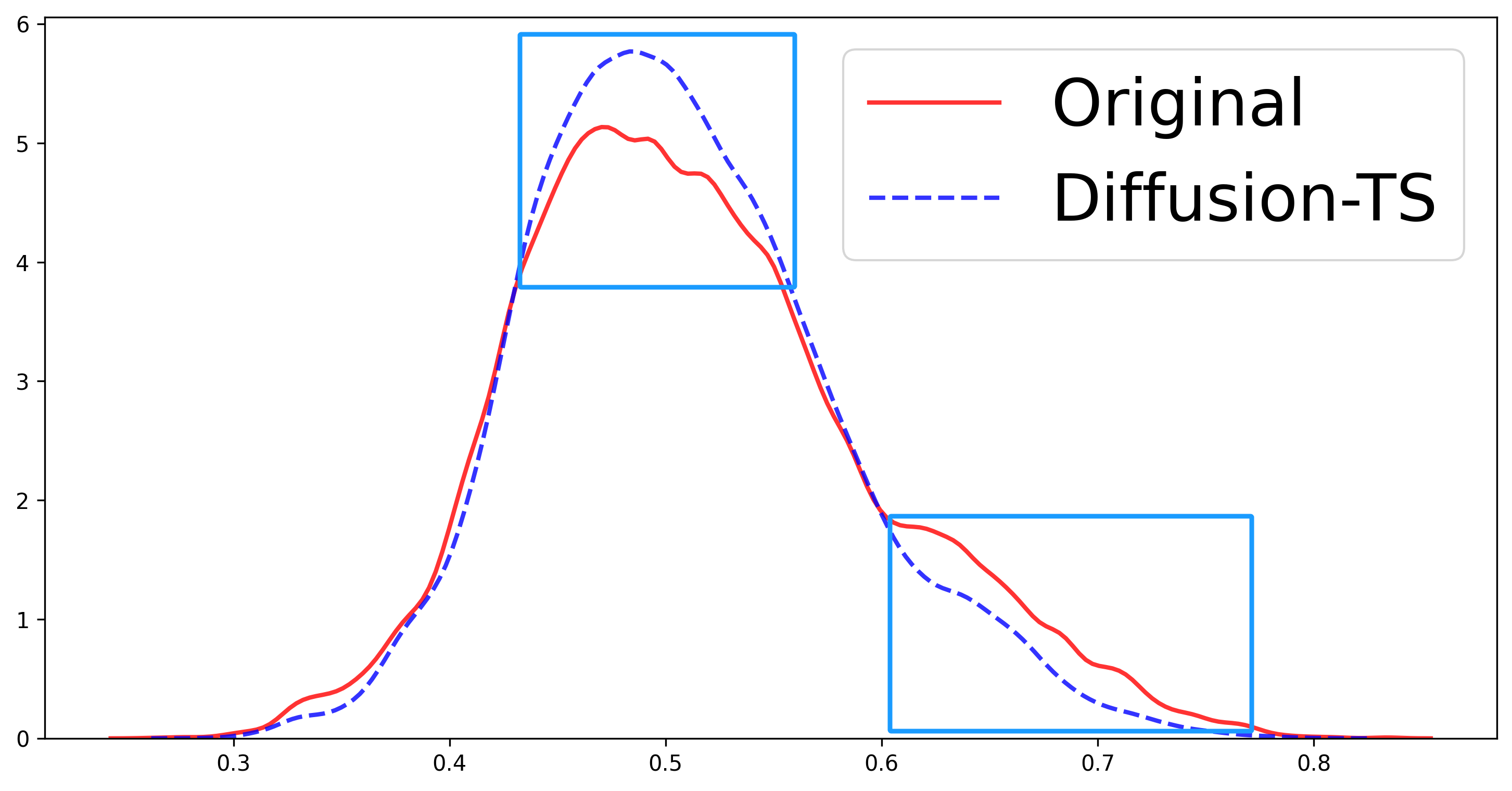}
        \caption*{(c) \texttt{ETTh}-96}
    \end{minipage}%
    \quad
    \begin{minipage}[b]{0.35\textwidth}
        \centering
        \includegraphics[width=\textwidth, trim=0 0 0 0, clip]{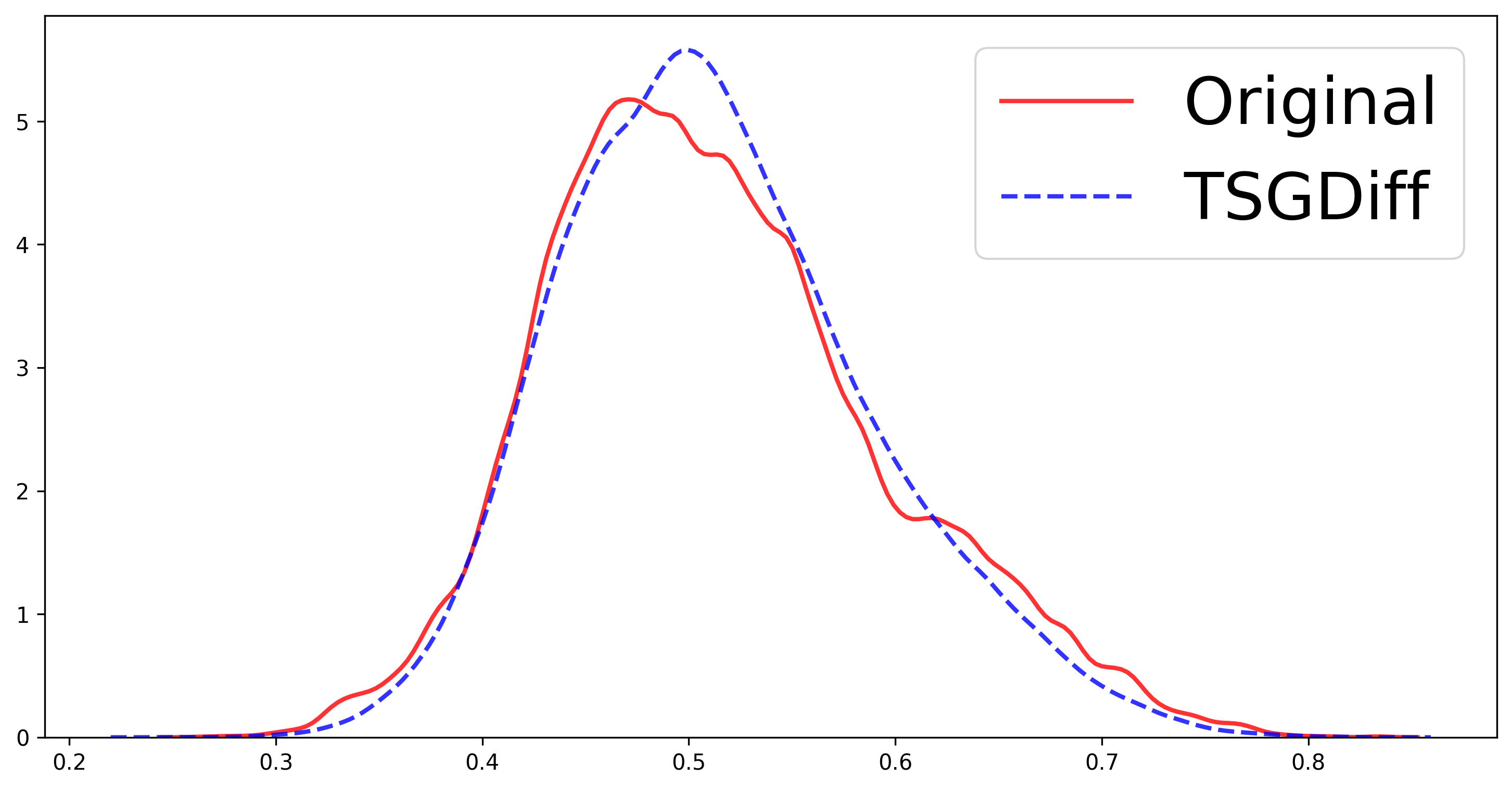}
        \includegraphics[width=\textwidth, trim=0 0 0 0, clip]{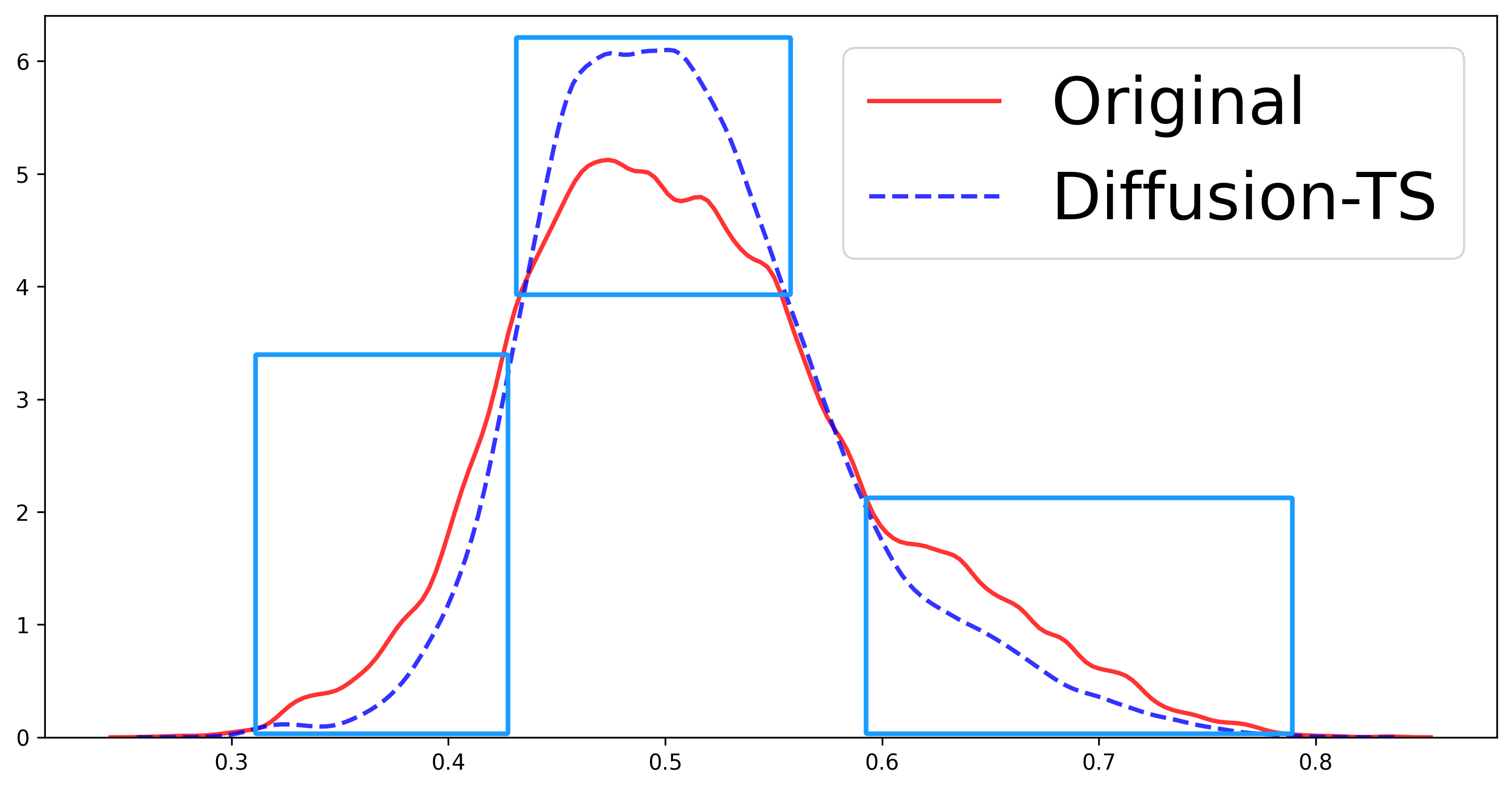}
        \caption*{(d) \texttt{ETTh}-128}
    \end{minipage}
    \caption{kernel density estimation plots of the time series of length 24, 64, 128 and 256 synthesized by \textit{TSGDiff} and Diffusion-TS. Red dots represent real data instances, and blue dots represent generated data samples in all plots.}
    \label{fig:kernellong}
\end{figure*}

\textbf{Kernel Density Estimation:}
Figure \ref{fig:kernellong} shows the kernel density estimation results for time series of lengths 48, 64, 96 and 128. Both \textit{TSGDiff} and Diffusion-TS produce distributions nearly identical to the original data. However, as sequence length increases, \textit{TSGDiff} continues to generate synthetic data with a distribution highly similar to the original, while Diffusion-TS struggles to achieve the same level of consistency.

\end{document}